\documentclass{article}

\PassOptionsToPackage{numbers, compress}{natbib}

\usepackage[final]{neurips_2021}

\usepackage[utf8]{inputenc} % allow utf-8 input
\usepackage[T1]{fontenc}    % use 8-bit T1 fonts
\usepackage{hyperref}       % hyperlinks
\usepackage{url}            % simple URL typesetting
\usepackage{booktabs}       % professional-quality tables
\usepackage{xcolor}         % colors
\usepackage{amsmath}
\usepackage{amsthm}
\usepackage{amsfonts}       % blackboard math symbols
\usepackage{amssymb}
\usepackage{stmaryrd}
\usepackage{nicefrac}       % compact symbols for 1/2, etc.
\usepackage{microtype}      % microtypography
\usepackage{subfigure}
\usepackage{enumerate}
\usepackage{wrapfig}
\usepackage[framemethod=default]{mdframed}
\usepackage{graphicx}
\usepackage{balance}
\usepackage{paralist}
\usepackage[normalem]{ulem}

\newtheorem{theo}{Theorem}
\newtheorem{prop}{Proposition}
\newtheorem{lemm}{Lemma}
\newtheorem{proper}{Property}

\title{Medical Dead-ends and Learning to Identify High-risk States and Treatments}

\author{%
  Mehdi Fatemi\\
  Microsoft Research\\
  \texttt{mehdi.fatemi@microsoft.com} \\
   \And
   Taylor W. Killian\\
   University of Toronto, Vector Institute\\
   \texttt{twkillian@cs.toronto.edu} \\
   \AND
   Jayakumar Subramanian \\
   Media and Data Science Research, Adobe India \\
   \texttt{jayakumar.subramanian@gmail.com} \\
   \And
   Marzyeh Ghassemi \\
   Massachusetts Institute of Technology \\
   \texttt{mghassem@mit.edu} \\
}

\begin{document}

\maketitle

\begin{abstract}
 Machine learning has successfully framed many sequential decision making problems as either supervised prediction, or optimal decision-making policy identification via reinforcement learning. In data-constrained offline settings, both approaches may fail as they assume fully optimal behavior or rely on exploring alternatives that may not exist. We introduce an inherently different approach that identifies possible ``dead-ends'' of a state space. We focus on the condition of patients in the intensive care unit, where a ``medical dead-end'' indicates that a patient will expire, regardless of all potential future treatment sequences. We postulate ``treatment security'' as avoiding treatments with probability proportional to their chance of leading to dead-ends, present a formal proof, and frame discovery as an RL problem. We then train three independent deep neural models for automated state construction, dead-end discovery and confirmation. Our empirical results discover that dead-ends exist in real clinical data among septic patients, and further reveal gaps between secure treatments and those that were administered.
\end{abstract}

\section{Introduction}
Off-policy Reinforcement Learning (RL) was designed as the way to isolate behavioural policies, which generate experience, from the target policy, which aims for optimality. It also enables learning multiple target policies with different goals from the same data-stream or from previously recorded experience~\cite{sutton_book}. This algorithmic approach is of particular importance in safety-critical domains such as robotics~\cite{kober2013reinforcement}, education~\cite{mandel2014offline} or healthcare~\cite{murphy2001marginal} where data collection should be regulated as it is expensive or carries significant risk. Despite significant advances made possible by off-policy RL combined with deep neural networks~\cite{mnih2015human,lillicrap2015continuous, pmlr-v80-espeholt18a}, the performance of these algorithms degrade drastically in fully \textit{offline} settings \citep{lange2012batch}, without additional interactions with the environment~\cite{jaques2019way,fujimoto2019off}.
These challenges are deeply amplified when the dataset is limited and exploratory new data cannot be collected for ethical or safety purposes. This is because robust identification of an optimal policy requires exhaustive trial and error of various courses of actions~\cite{bertsekas_neuro, kushner2003}. In such fully offline cases, naively learned policies may significantly overfit to data-collection artifacts~\cite{francois2019overfitting, sinha2021s4rl, agarwal2020optimistic}. Estimation errors due to limited data may further lead to mistimed or inappropriate decisions with adverse safety consequences~\citep{rebba2006validation}.

Even if optimality is not attainable in such constrained cases, negative outcomes in data can be used to identify behaviors to avoid, thereby guarding against overoptimistic decisions in safety-critical domains that may be significantly biased due to reduced data availability. In one such domain, healthcare, RL has been used to identify optimal treatment policies based on observed outcomes of past treatments~\cite{yu2019reinforcement}. These policies correspond to advising \emph{what treatments to administer}, given a patient's condition. Unfortunately, exploration of potential courses of treatment is not possible in most  clinical settings due to legal and ethical implications; hence, RL estimates of optimal policies are largely unreliable in healthcare~\cite{Gottesman2019}. 

In this paper, we develop a novel RL-based method, Dead-end Discovery (DeD), to identify \emph{treatments to avoid} as opposed to what treatment to select. Our paradigm shift avoids pitfalls that may arise from constraining policies to remain close to possibly suboptimal recorded behavior as is typical in current state of the art offline RL approaches~\cite{fujimoto2019off,kumar2019stabilizing,wu2019behavior,wang2020critic}. When the data lacks sufficient amounts of exploratory behavior, these methods fail to attain a reliable policy. We instead use this data to constrain the scope of the policy, based on retrospective analysis of observed outcomes, a more tractable approach when data is limited. Our goal is \textit{to avoid future dead-ends} or regions in the state space from which negative outcomes are inevitable (formally defined in Section~\ref{sec:special_states}). DeD identifies dead-ends via two complementary Markov Decision Processes (MDPs) with a specific reward design so that the underlying value functions will carry special meaning (Section~\ref{sec:method}). These value functions are independently estimated using Deep Q-Networks (DQN)~\cite{mnih2015human} to infer the likelihood of a negative outcome occurring (D-Network) and the reachability of a positive outcome (R-Network). Altogether DeD formally connects the notion of value functions to the dead-end problem, learned directly from offline data.

We validate DeD in a carefully constructed toy domain, and then evaluate real health records of septic patients in an intensive care unit (ICU) setting~\cite{JohnsonPollardShenEtAl2016}. Sepsis treatment and onset is a common task in medical RL~\cite{henry2015targeted,futoma2017improved,komorowski2018artificial,saria2018individualized} because the condition is highly prevalent~\cite{DEUTSCHMAN2014463, Singer_2016}, physiologically severe~\cite{VINCENT2013774}, costly~\cite{hcup:2013} and poorly understood~\cite{martin2012sepsis}. Notably, the treatment of sepsis itself may also contribute to a patient's deterioration~\cite{pmid28130687,pmid25072761}, thus making treatment avoidance a particularly well-suited objective. We find that DeD confirms the existence of dead-ends and demonstrate that 12\% of treatments administered to terminally ill patients reduce their chances of survival, some occurring as early as 24 hours prior to death. The estimated value functions underlying DeD are able to capture significant deterioration in patient health 4 to 8 hours ahead of observed clinical interventions, and that higher-risk treatments possibly account for this delay. Early identification of suboptimal treatment options is of great importance since sepsis treatment has shown multiple interventions within tight time frames (10 to 180 minutes) after suspected onset decreases sepsis mortality~\cite{gauer2020sepsis}. 

While motivated by healthcare, we propose the use of DeD in safety-critical applications of RL in most data-constrained settings. We introduce a formal methodology that outlines how DeD can be implemented within an RL framework for use with real-world offline data.
We construct and train DeD in a generic manner which can readily be used for other data-constrained sequential decision-making problems. In particular, we emphasize that DeD is well suited to analyze high-risk decisions in real-world domains. 

\vspace{-0.25cm}

\section{Related Work}

{\raggedright\textbf{RL in Health:}\setlength{\parindent}{15pt}}
RL has been the subject of much focus in health~\cite{yu2019reinforcement}, with particular emphasis on sepsis seeking to develop optimal treatment recommendation policies~\cite{henry2015targeted,futoma2017improved,komorowski2018artificial,saria2018individualized,raghu2017continuous,peng2018improving,li2019optimizing,tang2020clinician}. However, with fixed retrospective medical data, an optimal policy that maximizes a patient's chance of recovery is both computationally and experimentally infeasible.
To our knowledge, we are the first to target improved treatment recommendations by avoiding high-risk treatments in a fully offline manner.

{\raggedright\textbf{Safety in RL:}\setlength{\parindent}{15pt}}
RL has a rich history in safety~\cite{garcia2015comprehensive}, with recent work attempting to limit high risk actions by constraining parametric uncertainty~\cite{thomas2015safe}, through alignment between agent and human objectives~\cite{hadfield2016cooperative,hadfield2017inverse}, by directly constraining the agent optimization process to avoid unsafe actions~\cite{thomas2019preventing}, or by improving over a baseline policy~\cite{laroche2019safe}. In these settings model performance is evaluated in online settings where more data can be acquired or models can be tested against new cases as well as known baselines. We focus on the more challenging offline setting with limited and non-exploratory data, reflecting the reality of healthcare settings.

{\raggedright\textbf{Dead-ends:}\setlength{\parindent}{15pt}}
The concept of dead-ends and the corresponding security condition that we build from was proposed by \citet{fatemi2019dead} in the context of \textit{exploration}. In their work an online RL agent needs to experience various courses of actions from each state, through which it learns optimal behavior. 
We adapt this approach and expand the theoretical results to an offline RL setting as is found within healthcare--where exploration is untenable--to determine which treatments increase the likelihood of entering a dead-end, based on the patient's current health state.

Related concepts to dead-ends were introduced by \citet{irpan2019off}, focused primarily on policy evaluation. The authors introduce a notion of \emph{feasible} states as those that are not \emph{catastrophic} and from which an agent will not immediately fail. Whether or not a state is feasible is determined via positive-unlabeled classification. This inherently differs from our approach where we formally characterize dead-ends and a corresponding security condition through which we can identify treatments to avoid that likely lead to dead-ends\footnote{A more in depth discussion on the differences between \citet{irpan2019off} and this work can be found in Appendix Section~\ref{appendix:sec:related}}. Our formalization is discussed in the next section.

\section{Methods}
\label{sec:problem_construction}

\subsection{Preliminaries} Our pipeline isolates state construction from value estimation with RL. Therefore we consider episodic Markov Decision Processes (MDP) $\mathcal{M} = (\mathcal{S}, \mathcal{A}, T, r, \gamma)$, where $\mathcal{S}$ and $\mathcal{A}$ are the discrete sets of states and treatments\footnote{Our results can easily be extended to continuous state-spaces by properly replacing summations with integrals. For brevity, we only present formal proofs for the discrete case. Additionally, as our primary motivating domain lies within healthcare, we use the term ``treatment'' in place of ``action''.}; $T: ~ \mathcal{S}\times \mathcal{A} \times \mathcal{S} \rightarrow [0, 1]$ is a function that defines the probability of transitioning from state $s_t$ to $s_{t+1}$ if treatment $a_t$ is administered; $R: \mathcal{S}\times \mathcal{A} \times \mathcal{S} \rightarrow [r_{min}, r_{max}]$ is a finite reward function and $\gamma\in [0, 1]$ denotes a scalar discount factor. 

A \emph{policy} $\pi(s, a)=\mathbb{P}[A_{t}=a|S_{t}=s]$ defines how treatments are selected, given a state. A \emph{trajectory} is comprised of sequences of tuples $(S_t, A_t, R_t, S_{t+1})$ with $S_0$ being the initial state of the trajectory. Sequential application of the policy is used to construct trajectories. 
The reward collected over the course of a trajectory induces the \emph{return} $G_{t}=\sum_{j=0}^{\infty}\gamma^{j}R_{t+j+1}$. We assume that all the returns are finite and bounded. A trajectory is considered \emph{optimal} if its return is maximized. A state-treatment value function $Q^{\pi}(s,a)=\mathbb{E}^{\pi}[G_{0}| S_{0}=s, A_{0}=a]$ is defined in conjunction with a policy $\pi$ to evaluate the expected return of administering treatment $a$ at state $s$ and following $\pi$ thereafter. The optimal state-treatment value function is defined as $Q^{*}(s,a)=\max_{\pi} Q^{\pi}(s,a)$, which is the maximum expected return of all trajectories starting from $(s,a)$. We define state value and optimal state value as $V^{\pi}(s) = \mathbb{E}_{a\sim \pi} Q^{\pi}(s,a)$ and $V^{*}(s) = \max_{a} Q^{*}(s,a)$.

\subsection{Special States}
\label{sec:special_states}
We define a terminal state as the final observation of any recorded trajectory. We focus on two types of terminal state that correspond to positive or negative outcomes. 
Our goal is to identify all \emph{dead-end} states, from which negative outcomes are unavoidable (happening w.p.1), regardless of future treatments.
In safety-critical domains, it is crucial to avoid such states \emph{and} identify the probability with which any available treatment will lead to a dead-end. 
We also introduce the complementary concept of \emph{rescue} states, from which a positive outcome is \emph{reachable} with probability one. If an agent is in a rescue state, there exists at least one treatment at each time step afterwards which leads to either another rescue state or the eventual positive outcome.
The fundamental contrast between dead-end and rescue states is that if the agent enters a rescue state, it \emph{does not} mean the treatment process is done; it rather means that at each time step afterwards there exists at least one treatment to be found and administered until the positive outcome occurs. There might be trajectories starting from a rescue state which include non-rescue states. This is not the case for a dead-end state.

Formally, we augment $\mathcal{M}$ with a non-empty termination set $\mathcal{S}_{T}\subset \mathcal{S}$, which is the set of all terminal states. Mathematically, a terminal state is absorbing (self-transition w.p.1) with zero reward afterwards. All terminal states are by definition zero-valued, but the transitions to them may be associated with a non-zero reward. We require that, from all states, there exists at least one trajectory with non-zero probability arriving at a terminal state. In an offline setting with limited and non-exploratory data, inducing an optimal policy \textit{is not feasible}~\cite{kushner2003}; hence, we do not specify the reward function of $\mathcal{M}$ for which standard RL would optimize cumulative rewards, but in later sections present a specific design of reward (and discount factor) to assist in identifying dead-end/rescue states.
Finally, the sets of dead-end and rescue states are denoted respectively by $\mathcal{S}_{D}$ and $\mathcal{S}_{R}$. We formally distinguish dead-end/rescue states from the outcome, asserting that $\mathcal{S}_{D},\mathcal{S}_{R}\not\subset\mathcal{S}_{T}$.

\subsection{Treatment Security}
When dealing with data-constrained offline scenarios, a core distinction is necessary: Realization of an optimal treatment at a given state requires knowledge of all future outcomes for all possible treatments, which is not feasible. However, the data may contain enough information to estimate the possible outcome of a certain treatment at a similar state. If such an outcome is negative with high probability, then we should advise against the treatment, even if an optimal treatment still remains unknown. This distinction leads to a paradigm shift from finding the best possible treatment to mindful avoidance of dangerous ones. This shift further motivates a different design space to make use of the limited, yet available data. 

We adapt the security condition from \citet{fatemi2019dead} and formalize the treatment avoidance problem with a more generalized \emph{treatment security condition}. We note that the chance of a negative outcome is best described by the probability of falling into a dead-end or immediate negative termination. The security condition therefore constrains the scope of a given behavioral policy $\pi$ if \emph{any} knowledge exists about dead-ends or negative termination. Formally, if at state $s$, treatment $a$ leads to a dead-end with probability $P_{D}(s,a)$ or immediate negative termination with probability $F_{D}(s,a)$ with a level of certainty $\lambda\in[0,1]$, then $\pi$ must avoid selecting $a$ at $s$ with the same certainty:
\begin{align} \label{eq:security}
P_{D}(s, a) + F_{D}(s, a)  \geq \lambda ~ \Longrightarrow ~  \pi(s,a) \le 1 - \lambda.
\end{align}
E.g., if a treatment leads to a dead-end or termination with probability more than $80\%$, then that treatment should be selected for administration no more than $20\%$ of the time. While we would like \eqref{eq:security} to hold for the maximum $\lambda$, inferring such maximal values is intractable for all state-treatment pairs. Moreover, directly computing $P_{D}$ and $F_{D}$ would require explicit knowledge of all dead-end and negative terminal states as well as all transition probabilities for future states. These make the application of \eqref{eq:security} nearly impossible. We next develop a learning paradigm to enable \eqref{eq:security} from data. 

\subsection{Dead-end Discovery (DeD)} \label{sec:method}

In order to identify and confirm the existence of dead-end states, we construct two Markov Decision Processes (MDPs) $\mathcal{M}_D$ and $\mathcal{M}_R$ to be identical to $\mathcal{M}$, with $\gamma = 1$ for both. We also define the following reward functions: $\mathcal{M}_{D}$ returns $-1$ with any transition to a negative terminal state (and zero with all other transitions) whereas $\mathcal{M}_{R}$ returns $+1$ with any transition to a positive terminal state (zero otherwise). Let $Q^{*}_{D}$, $Q^{*}_{R}$, $V^{*}_{D}$ and $V^{*}_{R}$ denote the optimal state-treatment and state value functions of $\mathcal{M}_{D}$ and $\mathcal{M}_{R}$, respectively. Note that due to the reward functions of these MDPs, for all states and treatments, $Q^{*}_{D}(s,a)\in [-1,0]$ and $Q^{*}_{R}(s,a)\in[0,1]$.

Having selected treatment $a$ at state $s$, using the Bellman equation, we prove\footnote{All proofs to the theoretical claims presented in this paper can be found in Appendix \ref{appendix:sec:proofs}} that 
\begin{align} \label{eq:qd}
-Q_{D}^{*}(s,a) = P_{D}(s,a) + F_{D}(s,a) + M_{D}(s,a)
\end{align}
In addition to the quantities defined previously, $M_{D}(s,a)$ denotes the probability of circumstances in stochastic environments where a negative terminal state ultimately occurs despite receiving optimal treatments at all steps in the future. Equation \eqref{eq:qd} therefore reveals that $-Q_{D}^{*}$ carries special physical meaning: it corresponds to the \emph{minimum probability of a negative outcome}, because future treatments may not necessarily be optimal.
Equivalently, $1 + Q_{D}^{*}(s,a)$ can be seen as the \emph{maximum hope of a positive outcome} if treatment $a$ is administered at state $s$.

Building from \citet{fatemi2019dead}, we show that $V_{D}^{*}$ of all dead-end states %---and only dead-end states---
will be precisely $-1$. %(Appendix \ref{appendix:sec:proofs}, Lemma 1). 
By extension, $Q_{D}^{*}(s,a)=-1$ for all treatments $a$ at state $s$ if and only if $s$ is a dead-end. In fact, $1 + Q^{*}_{D}(s, a)$ provides an appropriate threshold to secure any given policy $\pi(s, a)$. More formally, the following statement guarantees treatment security as presented in \eqref{eq:security} for all values of $\lambda$:
\begin{align} \label{eq:main-result1}
    \pi(s, a) \le 1 + Q^{*}_{D}(s, a)
\end{align}
In short, for treatment security it is sufficient to abide by the maximum hope of a positive outcome. This construction directly connects the RL concept of value functions to dead-end discovery. While $V_{D}^{*}(s)$ enables detecting dead-end states, \eqref{eq:main-result1} leverages $Q_{D}^{*}$ for treatment avoidance. We establish parallel results for rescue states similarly. The following theorem summarizes the theory and shapes the basis of DeD. See Appendix~\ref{appendix:sec:proofs} for the proof and further details.

{\raggedright\textbf{Theorem 1}\setlength{\parindent}{15pt}.} Let treatment $a$ be administered at state $s$, and $P_{D}(s, a)$ and $P_{R}(s, a)$ denote the probability of transitioning to a dead-end or rescue state. Similarly, let $F_{D}(s, a)$ and $F_{R}(s, a)$ denote the probability of transitioning to either a negative or positive terminal state. The following hold:
\begin{enumerate}[T1]
    \item $P_{D}(s,a)+F_{D}(s,a)=1$ if and only if $Q^{*}_{D}(s,a)=-1$.
    \item $P_{R}(s,a)+F_{R}(s,a)=1$ if and only if $Q^{*}_{R}(s,a)=1$.
    \item There exists a threshold $\delta_{D}\in(-1,~0)$ independent of states and treatments, such that $Q^{*}_{D}(s,a)\ge \delta_{D}$ for all $s$ and $a$, unless $P_{D}(s,a)+F_{D}(s,a) = 1$.
    \item There exists a threshold $\delta_{R}\in(0,~1)$ independent of states and treatments, such that $Q^{*}_{R}(s,a)\le \delta_{R}$ for all $s$ and $a$, unless $P_{R}(s,a)+F_{R}(s,a) = 1$.
    \item For any policy $\pi$, state $s$, and treatment $a$, if $\pi(s,a) \le 1+Q^{*}_{D}(s,a)$ and $\lambda\in [0,1]$ exists such that $P_{D}(s,a)+F_{D}(s,a) \ge \lambda$, then $\pi(s,a) \le 1-\lambda$.
    \item For any policy $\pi$, state $s$, and treatment $a$, if $\pi(s,a) \ge Q^{*}_{R}(s,a)$ and $\lambda\in [0,1]$ exists such that $P_{R}(s,a)+F_{R}(s,a) \ge \lambda$, then $\pi(s,a) \ge \lambda$.
\end{enumerate}
It is immediate from (T1) and (T2) that $Q^{*}_{D}$ and $Q^{*}_{R}$ incorporate complete information when transitioning to a dead-end state or to a rescue state as a result of administrating treatment $a$ at $s$. (T3) assures that a threshold $\delta_{D}$ exists to separate treatments that lead immediately to dead-ends from alternatives. 
(T4) allows us to confirm a dead-end by examining if $Q^{*}_{R}$ is also smaller than some threshold $\delta_{R}$. No dead-end can violate $\delta_{R}$ due to (T4) and such a threshold exists. If $Q^{*}_{D}$ is available and $\delta_{D}$ is known, then this step is redundant. However, without access to $Q^{*}_{D}$ and an accurate $\delta_{D}$, (T4) helps to confirm any presumed dead-end. Finally, (T5) provides the means by which the treatment policy is guided to avoid dangerous treatments. (T6) is used to also confirm whether the treatment should be avoided. We explain how to practically select the thresholds $\delta_{D}$ and $\delta_{R}$ in Sec. \ref{sec:results}. 

Of note, by definition, value functions encompass long-term consequences and are not myopic to possible immediate events, as opposed to supervised learning from immediate observation of an outcome. This inherent characteristic of value functions indeed yields the theoretical result presented by Lemma 2 (Appendix Sec. A1), one result of which is that $-Q_D$ corresponds to the minimum probability of a negative future outcome. Supervised learning from immediate outcomes, on the other hand, lacks this formal property~\cite{maley2020mortality}; hence, it is not expected to provide parallel results with DeD.

\subsection{Neural Network Based State Construction and Identification}
{\flushleft\textbf{State construction (SC-Network).}} In domains where solitary observations do not carry salient information for learning the decision-making process, states may need to be constructed from data using a neural network. In these circumstances a separate SC-Network can be used to transform a single or possible sequence of observations into a fixed embedding, considered the state $s$ at time $t$.

{\flushleft\textbf{Identification (D-Network and R-Network).}} In order to approximate $Q^{*}_{D}$ and $Q^{*}_{R}$, two separate neural networks can be used to compute $Q_{D}$ and $Q_{R}$ for all treatments given a state constructed by the SC-Network. With trained $Q_{D}$ and $Q_{R}$ networks, we can then apply thresholds $\delta_{D}$ and $\delta_{R}$ as specified in Theorem 1.
As data is limited and non-exploratory, approximation error is inevitable. To mitigate this limitation, the method's sensitivity can be adjusted by adapting the thresholds $\delta_{D}$ and $\delta_{R}$ (additionally, see Proposition 1 and Remarks 1-4 in Appendix \ref{appendix:sec:proofs}). Smaller thresholds result in more false negative and less false positive cases. Of note, value-overestimation, a known limitation of deep RL models, will often cause $Q_{D}$ and $Q_{R}$ to be larger than $Q^{*}_{D}$ and $Q^{*}_{R}$ respectively. This naturally reduces false positives while increasing false negatives.

\subsection{Toy Problem Validation: Life-Gate}

\begin{figure}[t!]
    \centering
    \includegraphics{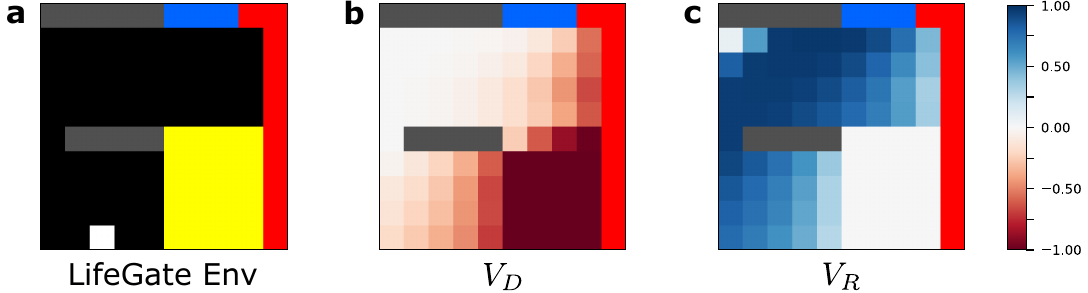}
    \caption{\textbf{The Life-Gate Example.} The tabular navigation task of life-gate is illustrated in (\textbf{a}). Corresponding dead-end and rescue state-value functions, $V_{D}$ and $V_{R}$, are shown in (\textbf{b}) and (\textbf{c}). The value functions are learned through Q-learning and with the definition of $\mathcal{M}_D$ and $\mathcal{M}_R$.}
    \label{fig:lifegate_exp_maintext}
\end{figure}

We briefly provide a tabular toy-example (Life-Gate), which involves dead-ends, to empirically illustrates the merit of Theorem 1 by learning $Q^{*}_{D}$ and $Q^{*}_{R}$ (Figure \ref{fig:lifegate_exp_maintext}). This toy set-up comprises an interesting case, where the agent faces an environment to examine with no knowledge of possible dangers. Importantly, once a dead-end state (yellow) is reached, it may take some random number of steps before reaching a ``death gate'' (red). All along such trajectories of dead-end states, the agent still has to choose actions with the (false) hope of reaching a ``life-gate'' (blue). Discovering any single dead-end state and signaling the agent when it is approached would be of significant importance. On the other hand, adjacent states to dead-ends are possibly the most critical to alert, as it might be the last chance to still do something to avoid failure (see Appendix \ref{sec:lifegate_apdx} for more details). 

We next use the tools provided by Theorem 1. 
The value functions are more than 90\% trained, still allowing learning errors. 
In this example, even with the errors due to lack of full convergence, $\delta_D = -0.7$ and $\delta_R = 0.7$ seem to clearly set the boundary for most states (with a few exceptions due to the errors). If a state is observed whose $V_D$ and $V_R$ values violate these thresholds, the state can be flagged as a dead-end with high probability. Setting a lower threshold can help to raise flags earlier on, when the conditions are of high-risk, but it is still not too late. We can apply the same thresholds to further flag high-risk actions (not shown). Lastly, we note (T1) from Theorem 1. It can be seen that only for all the yellow area (aside from the few erroneous states), $V_D=-1$. Clearly, no dead-end state can be a rescue, as seen by $V_R=0$ for the yellow area too.

\section{Empirical Setup for Dead-end Analysis} \label{sec:setup}

{\flushleft\textbf{Data: }}We use DeD to identify medical dead-ends in a cohort of septic patients drawn from the MIMIC (Medical Information Mart for Intensive Care) - III dataset (v1.4)~\cite{JohnsonPollardShenEtAl2016,mimicweb}. This cohort totals 19,611 patients (17,730 survivors and 1,881 nonsurvivors), with 44 observation variables, and 25 treatment choices (5 discrete levels for each of IV fluid and vasopressor). We follow prior work~\cite{komorowski2018artificial} and aggregate each variable every four hours using the per-patient variable mean if data is present, or impute using the value from the nearest neighbor. 

{\flushleft\textbf{Terminal States.}} In our ICU setting, possible terminal states are either patient recovery (discharge from ICU) or death. We define ``death'' as the last recorded point in the EMR of nonsurviving patients when expiration is imminent, but may not necessarily be the biological point of death. In practice this definition of terminal state may occur hours or days before biological death and covers situations where care support devices are disconnected, when a patient requests a cessation of treatment, etc.

Our goal is to identify all \emph{medical dead-end} states, defined as patient states from which death is unavoidable, regardless of future treatments. Relatedly, we also desire to discover all treatments that may possibly lead to a medical dead-end state in order to learn which treatments to avoid.

{\flushleft\textbf{SC-Network.}} 
As observations of patient health are inherently partial, we need an informative latent representation of state~\cite{killian2020empirical}, sufficient for evaluating treatment security. To form these state representations we process a sequence of observations prior to and including any time $t$ as well as the last selected treatment to form the state $s_t$.
We train a standalone State Construction (SC) network using Approximate Information State (AIS)~\cite{subramanian2019approximate} in a self-supervised manner for this purpose. Details of AIS and how it is used to train the SC-Network are included in Appendix \ref{sec:state_const}.

{\flushleft\textbf{D-Network and R-Network.}} Computed states are given as input to the D- and R-networks to approximate $Q^{*}_{D}$ and $Q^{*}_{R}$. We use the double-DQN algorithm \cite{Hasselt2016} to train each network (details included in Appendix~\ref{appendix:sec:rl-training}). The outputs of trained D- and R- Networks produce value estimates of both the embedded patient state and all possible treatments to evaluate the probability of transitioning to a dead-end. This process of determining possibly high-risk treatments is central to DeD.

{\flushleft\textbf{Training: }}We train the SC-, D-, and R- networks in an offline manner, using retrospective data (\ref{fig:process}).
All models are trained with 75\% of the patient cohort (14,179 survivors, 1,509 nonsurvivors), validated with 5\% (890 survivors, 90 nonsurvivors), and we report all results on the remaining held out 20\% (2,660 survivors, 282 nonsurvivors). Further details of how the patient cohort is processed are provided in Appendix \ref{sec:data}.
Finally, to mitigate the data imbalance between surviving and non-surviving patients we use an additional data buffer that contains only \textit{the last transition} of nonsurvivors trajectories. Thus, a stratified minibatch of size 64 is constructed of 62 samples from the main data, augmented with 2 samples from this additional buffer, all selected uniformly. This same minibatch structure is used for training each of the three networks. For the training details see Appendix \ref{sec:state_const} and \ref{appendix:sec:rl-training}. 

\section{Empirical Results}\label{sec:results}

\begin{figure}[t!]
  \begin{minipage}[c]{0.45\textwidth}
    \includegraphics[width=\textwidth]{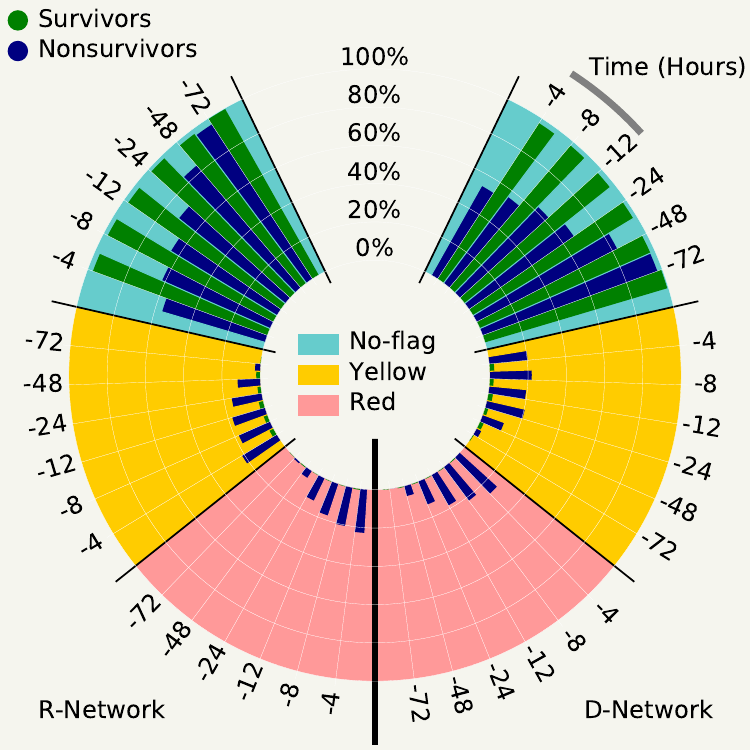}
  \end{minipage}
  \hspace*{\fill}
  \begin{minipage}[c]{0.46\textwidth}
    \caption{\textbf{Flag emergence for ICU patients.} Histograms of median $Q$ according to the flag status, for both surviving (green) and nonsurviving patients (blue) according to the R-Network (left) and D-Network (right). The bars are plotted according to the time prior to the recorded terminal state (the maximum trajectory length is 72 hours) and measure the percentage of patients whose states raise either a red, yellow or no flag. There is a clear worsening trend of state values for nonsurviving patients as they approach their terminal state, beginning as early as 48 hours prior.
    } \label{fig:hist_values}
    \hspace*{\fill}
  \end{minipage}
\end{figure}

\subsection{Septic Dead-End State Prediction}
{\raggedright\textbf{Experiment.}\setlength{\parindent}{15pt}}
In order to flag potentially non-secure treatments, we examine if $Q_{D}$ and $Q_{R}$ of each treatment at a given state pass certain thresholds $\delta_{D}$ and $\delta_{R}$, respectively. To flag potential dead-end states, we need to probe the state values, for which we examine the \textit{median} of $Q$ (rather than \textit{max} of $Q$) against similar thresholds. Using the median helps to avoid extreme approximation error due to generalization from potentially insufficient data. We found that $\delta_{D}=-0.25$ and $\delta_{R}=0.75$ minimize both false positives and false negatives, and use these as the thresholds for raising ``red'' flags. We also define a second, looser threshold of $\delta_{D}=-0.15$ and $\delta_{R}=0.85$, as raising ``yellow'' flags with higher sensitivity but increased false positives. This looser threshold targets an early indication of a patient's health condition deteriorating toward a dead-end state. In Appendix \ref{fig:hist_full_values} we report histogram of values at different quantiles, from which we established these thresholds.

{\raggedright\textbf{Results.}\setlength{\parindent}{15pt}}
Using the specified thresholds, DeD identifies increasing percentages of patients raising fatal flags as nonsurvivors approach death (Figure \ref{fig:hist_values} and Appendix \ref{table:prediction}). 
Note the distinctive difference between the trend of values in survivors (green bars) and nonsurvivors (blue bars). Over the course of 72 hours in the ICU, survivor trajectories raise nearly no red flag for both networks. In contrast, nonsurvivor trajectories demonstrate a steep reduction in \textit{no-flag} zone with increasing numbers of patients flagged in the \emph{Red} zone. The \emph{Yellow} zone is dominated largely by the nonsurvivors, yet there are also survivors who ultimately recover. 
Under the red-flag threshold, more than $12\%$ of treatments administered to non-surviving patients are identified to be detrimental 24 hours prior to death with a $0.6\%$ false positive rate (Appendix \ref{table:prediction}). We further identify that $2.7\%$ of non-surviving patient cases have entered unavoidable dead-end trajectories up to 48 hours before recorded expiration, with only a $0.2\%$ false positive rate, i.e., patients misidentified as near death. 

We find that 5\% of nonsurviving patients maintain the red flag for their last 24 hours recorded in the ICU before reaching a death terminal state. This monotonically increases to 13.9\% for patients who maintain a red flag through their final 8 hours of care (Appendix \ref{fig:duration}\textbf{b},\textbf{c}). These patients likely reached a dead-end with no subsequent chance of recovery; this is as compared to 89.3\% of nonsurviving test patients with no flag raised in their first 8 hours (Appendix \ref{fig:duration}\textbf{d}). 

There is a distinct difference between remaining on a flag for survivors and nonsurvivors (Appendix \ref{fig:duration}\textbf{a}). Even with our red threshold, very few survivors (0.5\%) raise and remain on red-flag for more than eight hours, decreasing to nearly zero for longer periods. In contrast, 32.6\% of nonsurvivors remain on red flags for similar duration with a fat tail. These results suggest that red-flag membership for long periods strongly correlates with mortality, inline with our theoretical analysis.

\begin{figure*}[t!]
\centering
\includegraphics[width=0.99\textwidth]{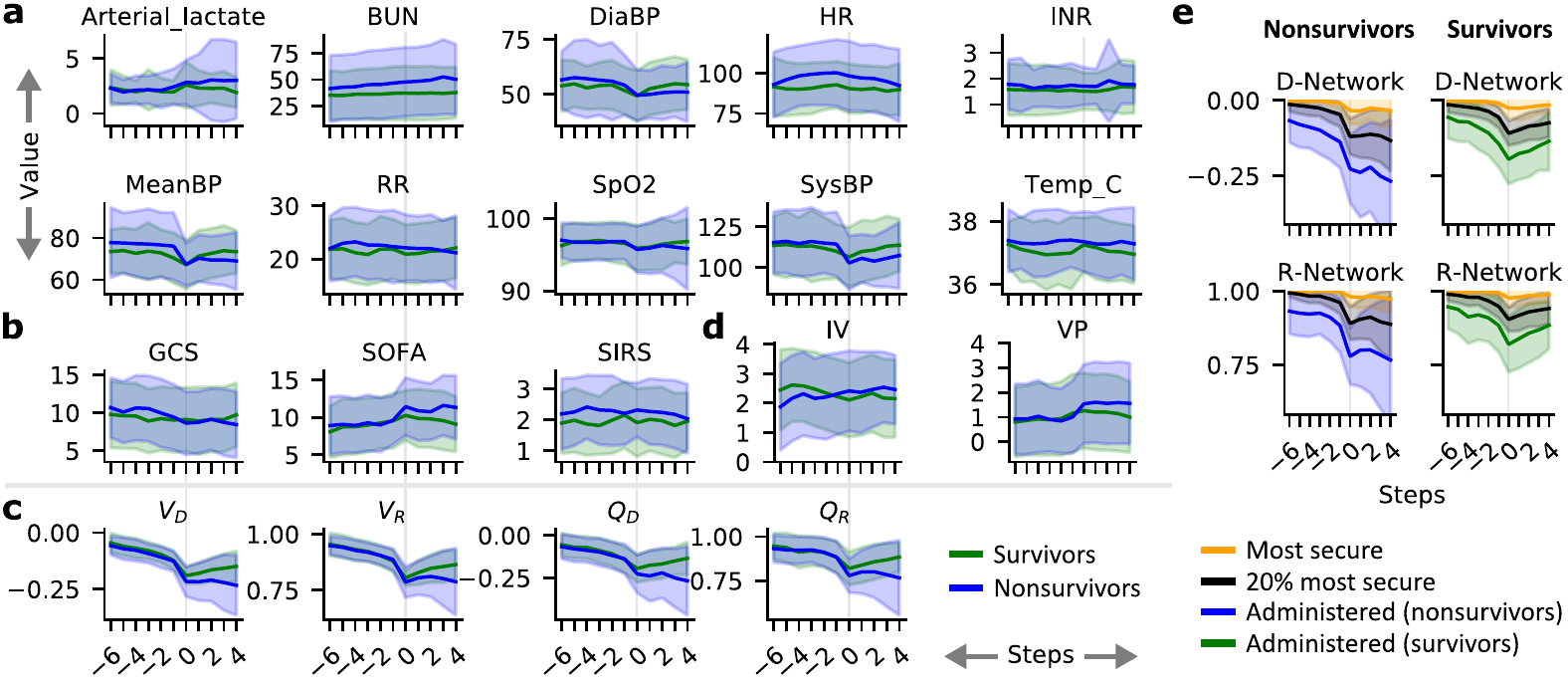}
\caption{\textbf{Trend of measures around the first raised flag.} Various measures are shown 24 hours (6 steps) before the first (red or yellow) flag is raised and 16 hours (4 steps) afterward. All nonsurviving (blue) and surviving (green) patient trajectories that fall within this window are averaged, shaded areas represent a single standard deviation. (\textbf{a}) selected key vital measures and lab tests, (\textbf{b}) established clinical measures, and (\textbf{c}) DeD value measures of state ($V$) and administered treatment ($Q$) from the D- and R-Networks and, (\textbf{d}) administered treatments. There is a clear turning point 4 to 8 hours prior to the flag being raised, which precisely corresponds to a drastic increase of VP and IV treatments. (\textbf{e}) the value of the maximum, the 5th maximum ($20\%$ best) and the actually administered treatment, demonstrating that better treatments were available when the chosen treatments were administered.
}
\label{fig:measures}
\end{figure*}

\subsection{First Flag Analysis}
{\raggedright\textbf{Experiment.}\setlength{\parindent}{15pt}}
To further support our hypothesis that dead-end states exist among septic patients and may be preventable, we align patients according to the point in their care when a flag is ``first raised''. We select all trajectories in the test data with at least 24 hours (6 steps) prior to the first flag and at least 16 hours (4 steps) afterwards (77 surviving and 74 nonsurviving patients). This window excludes patients with flags that occur either too early or too late. This allows for an investigation of the average trend of patient observations, administered treatments as well as the measures used in DeD over a sufficiently large window (Figure \ref{fig:measures}). 

{\raggedright\textbf{Results.}\setlength{\parindent}{15pt}}
The $V$ and $Q$ values estimated by DeD have similar behavior in survivors and nonsurvivors prior to the first flag, but values diverge after the flag is raised. Notably, the time step pinpointed by DeD to raise a flag corresponds to a similar diverging trend among various clinical measures, including SOFA and patient vitals (Figure~\ref{fig:measures}\textbf{a},\textbf{b}). This distinct behavior is also seen if looser threshold values are used for $\delta_{D}$ and $\delta_{R}$ (Appendix \ref{fig:measures_3flags}). After the flag is raised there is slight improvement in all value estimates, perhaps in response to the change in treatment. However the values of nonsurviving patient trajectories quickly collapse while survivors continue to improve.

The results of this analysis suggest two main points.  First, DeD identifies a clear \textit{critical point} in the care timeline where nonsurviving patients experiencing an irreversible deterioration in health. Second, there is a significant gap (Figure \ref{fig:measures}\textbf{e}) between the value of administered treatments and the 20\% most secure ones (5 out of 25). The critical point appears to arise when a patient's condition shifts towards improvement or otherwise enters a dead-end towards expiration. Perhaps most notable is that there is a clear inflection in the estimated values 4 to 8 hours prior to a flag being raised. Signaling this shift in the inferred patient response to treatment and the resulting flag may be used to provide an early indicator for clinicians (more conservative thresholds may be used to signal earlier). The trend of survivors shows that there is still hope to save the patient at this point. Note that \textit{all} these patients (survivors and nonsurvivors) are very similar in terms of both D/R values and their SOFA score prior to this point. This rejects the argument that survivors and nonsurvivors are inherently different. Additionally, while SOFA may appear correlated with DeD at the individual level, the trend of value functions can be noticeably more aggressive than SOFA with significantly less variance (\ref{fig:measures_3flags}). Further, most patients already have a high SOFA; hence, it is not sufficient for dead-end identification. DeD is however a provable methodology to this end.
Figures \ref{fig:measures}\textbf{d} and \textbf{e} advocate that the choice of treatment may play a role in entering dead-ends, since the divergence/drop occurs before the flag. The gap in value between the administered treatments and those with the highest estimated security suggest that better treatments were available, even for patients who eventually recover (Figures \ref{fig:measures}\textbf{e}).

\subsection{Individual Trajectories}
{\raggedright\textbf{Experiment.}\setlength{\parindent}{15pt}}
In our final analysis we extract relevant information surrounding a patient's value estimates from the electronic health record data, including the recorded clinical notes. We also use t-SNE~\cite{maaten2008visualizing} to project the state representations of the patient's trajectory, embedded using the SC-Network, among all recorded states in the test data (complete figures are presented in the Appendix). 

\begin{figure*}[t!]
\centering
\includegraphics[width=\textwidth]{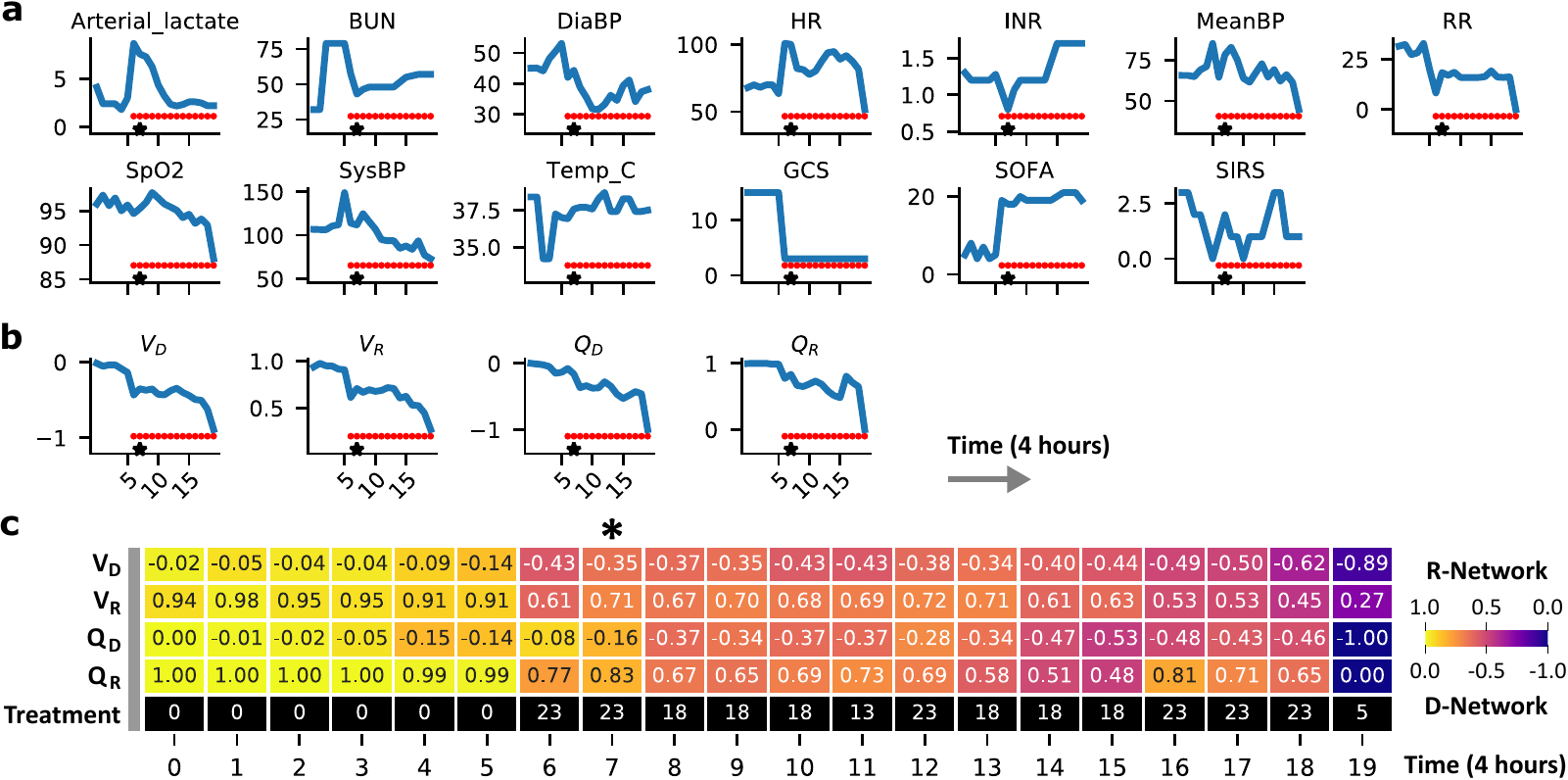}
\caption{\textbf{Events at ICU.} Certain vitals (\textbf{a}) and DeD value measures from both D- and R-Networks (\textbf{b}) are shown for a non-surviving patient (ICU-Stay-ID 262011). The black asterisks demonstrate the presumed onset of sepsis at step 7, and the color dots corresponds to the raised red or yellow flags. Lastly, (\textbf{c}) illustrates steps along the patient's trajectory with DeD estimated values %(colored correspondingly) 
and selected treatments. Notably, from steps 5 to 6 the state has a sudden jump to a low-value region that it fails to escape from, aligned with significant inflections in the recorded vitals, approximately 4 to 8 hours before presumed onset of sepsis. (See Appendix \ref{fig:full-traj_12139_icustayid_262011} -- \ref{fig:full-traj_6938_icustayid_235403} for full feature set plus accompanying excerpts from clinical notes of this and additional patients)}
\label{fig:tsne_notes}
\end{figure*}

{\raggedright\textbf{Results.}\setlength{\parindent}{15pt}}
The clinicians' chart notes confirm existence of dead-ends with a noted need for intubation, hypotension, and a discussion of moving the patient's care to ``comfort measures only'' (\ref{fig:full-traj_12139_icustayid_262011}\textbf{c}). Moreover, certain areas in the t-SNE projection of observed patient states appear to correspond with dead-end states (\ref{fig:full-traj_12139_icustayid_262011}\textbf{b}). Notably, the dramatic shift of clinically established measures such as SOFA and GCS closely follow the decrease in DeD estimated values (Figure \ref{fig:tsne_notes}\textbf{a}, \textbf{b}). This is similar to the trends seen prior to raised flags (Figure \ref{fig:measures}). This qualitative analysis suggests that the estimates of $Q_{D}$ and $Q_{R}$ are reliable and informative, supporting our prior conclusions.
Additional non-surviving patients are presented in Appendix~\ref{fig:full-traj_12139_icustayid_262011} -- \ref{fig:full-traj_6938_icustayid_235403}.

\section{Discussion}
In this work, we have introduced an RL-based approach for learning what treatments \emph{to avoid} based on observed patterns in limited offline data. We target avoiding treatments proportional to their chance of leading to dead-ends, regions of the state space from which negative outcomes are inevitable. We establish theoretical results that expand the concept of dead-ends in RL, facilitating the notification of high-risk treatments or, as applied to healthcare, septic patient conditions with increased likelihood of leading to a dead-end. Globally, sepsis is a leading cause of mortality \cite{DEUTSCHMAN2014463, VINCENT2014380, pmid26414292}, and an important end-stage to many conditions.
Consequently, even a slight decrease in mortality rate or improved efficacy of treatment could have a significant impact both in terms of saving lives and reducing costs.

Our work lays the groundwork for dead-end analysis in medical settings and is, to the best of our knowledge, the first use of RL to flag bad treatments rather than finding the best ones through estimating an optimal policy $\pi^{*}$. Our algorithm is generic, using RL methodology that is formally guaranteed to hold the security condition re-established in this paper. The discovery of dead-end states, and the treatments that likely lead to them, provides actionable insights in intensive care intervention. Further improvement of DeD's prediction quality could target additional features from the EMR environment, such as pre-ICU admission co-morbidities. In future work, we also hope to explore the specific drugs and dosages used in treatment.

Given its general construction, DeD is well matched for safety-critical applications of RL in data-constrained settings where it may be too expensive or unethical to collect additional exploratory data. With formal guarantees of satisfying the security condition, DeD is suitable for broader adoption when developing critical insights from retrospective data. Our framework is particularly relevant to data-constraint offline RL application domains such as robotics, industrial control, and automated dialogue generation where negative outcomes can be clearly identified~\cite{levine2020offline}.

{\raggedright\textbf{Limitations:}\setlength{\parindent}{15pt}}
While DeD is a promising framework for decision support in safety-critical domains with limited offline data, there are certain core limitations. While we use median values of $Q_D$ or $Q_R$ to avoid extreme extrapolation, training the D- and R- networks is still performed offline and extrapolation is likely still occurring. For simplicity we did not estimate $Q_D$ or $Q_R$ with contemporary offline RL methods; however, DeD is generic and replacing the DDQN learning method with more recent approaches would be straightforward, which can significantly improve the pipeline (we also note that finding $Q_D$ or $Q_R$ is an \textit{exponentially smaller} problem compared to finding $\pi^*$ to recommend best treatments). Additionally, we did not investigate the sensitivity of DeD to demographic information or with respect to specific features from the EMR. Thorough analysis of this sensitivity may elucidate the fairness and reliability of DeD. Finally, we did not externally validate DeD using data from a separate hospital or through investigation of suggested treatment avoidance by human clinicians. These investigations and more, concerning the causal entanglement of outcome and sequential treatments, are a focus of current and future work.

{\raggedright\textbf{Ethical Considerations and Societal Impact:}\setlength{\parindent}{15pt}}
This work, or derivatives of it, should never be used in isolation to exclude patients from being treated, e.g., not admitting patients or blindly ignore treatments. The treatment-avoidance part of our proposed approach is meant to shrink the scope of possible treatment options, and help the doctors make better decisions. Signalling high-risk states is also meant to warn the clinicians for immediate attention before it possibly becomes too late. In both cases, the flags that DeD supplies are statistically tied to the training data and unavoidable sources of error and bias and should not be seen as a binary treat/don’t treat decision. In particular, even in the case of red flags, the signals should not be interpreted as mathematical dead-ends with full precision. The intention of our approach is to assist clinicians by highlighting possibly unanticipated risks when making decisions and is not to be used as a stand-alone tool nor as a replacement of a human expert. Misuse of this algorithmic solution could carry significant risk to the well-being and survival of patients placed in a clinician’s care.

The primary goal of this work is to establish a proof of concept where especially high-risk treatments can be avoided, where possible, in context of a patient’s health condition. In acute care scenarios treatments come with inherent risk profiles and potential harms. In these settings tendencies to overtreat patients have arisen in attempt of ensuring their survival, increasing the chance of clinical errors to occur~\cite{carroll2017high}. Recent clinical research has sought to simplify practice to only the most necessary treatments~\footnote{see \url{http://jamanetwork.com/collection.aspx?categoryid=6017}}. In this spirit, we seek to infer the long-term impact of each available treatment in view of their risk of pushing the patient into a medical dead-end. The secondary goal of our work, on the other hand, is signal when the patient’s condition deteriorates, but may not be noticed by clinicians through monitoring clinical measures. This follows from the fact that DeD uses value functions, which provably enable such predictions.

\balance

\clearpage

\begin{ack}
We thank our many colleagues who contributed to thoughtful discussions and provided timely advice to improve this work. We specifically appreciate the feedback provided by Nathan Ng, Sindhu Gowda, and the RL team at MSR Montreal, as well as the encouragement and suggested improvements provided by anonymous reviewers.

This research was supported in part by Microsoft Research, a CIFAR Azrieli Global Scholar Chair, a Canada Research Council Chair, and an NSERC Discovery Grant.

Resources used in performing this research were provided, in part, by Microsoft Research, the Province of Ontario, the Government of Canada through CIFAR, and companies sponsoring the Vector Institute \url{www.vectorinstitute.ai/\#partners}.

\end{ack}

\section*{Data and Code Availability}
Our code and pretrained models to replicate the analysis (including figures) presented in this paper is located at \url{https://github.com/microsoft/med-deadend}. 

The MIMIC-III databases (DOI: 10.1038/sdata.2016.35) that support the findings of this study are publicly available through Physionet website: https://mimic.physionet.org, which facilitates reproducibility of the presented results. The cohort definition, extraction and preprocessing code can be found at \url{https://github.com/microsoft/mimic_sepsis}.

\section*{Author Contributions}
MF and MG designed the research. MF conceptualized theoretical ideas and developed formal results and proofs. JS developed the code for data generation and state construction and prepared initial experimental results. TK finalized the script to generate data from MIMIC, performed benchmarking of several state-construction algorithms--finalizing the decision to use AIS in the paper---and executed the experiments for the Life-Gate toy example. MF developed code for RL and made final experimental results. MF, TK and MG interpreted the results and wrote the manuscript. JS contributed to this work only during his internship at Microsoft Research. All the authors read and agreed on the final draft. 

\bibliography{ref}

\begin{thebibliography}{58}
\providecommand{\natexlab}[1]{#1}
\providecommand{\url}[1]{\texttt{#1}}
\expandafter\ifx\csname urlstyle\endcsname\relax
  \providecommand{\doi}[1]{doi: #1}\else
  \providecommand{\doi}{doi: \begingroup \urlstyle{rm}\Url}\fi

\bibitem[Sutton and Barto(1998)]{sutton_book}
R.~S. Sutton and A.~G. Barto.
\newblock \emph{Reinforcement Learning: An Introduction}.
\newblock MIT Press, 1998.
\newblock ISBN 0262193981.

\bibitem[Kober et~al.(2013)Kober, Bagnell, and Peters]{kober2013reinforcement}
Jens Kober, J~Andrew Bagnell, and Jan Peters.
\newblock Reinforcement learning in robotics: A survey.
\newblock \emph{The International Journal of Robotics Research}, 32\penalty0
  (11):\penalty0 1238--1274, 2013.

\bibitem[Mandel et~al.(2014)Mandel, Liu, Levine, Brunskill, and
  Popovic]{mandel2014offline}
Travis Mandel, Yun-En Liu, Sergey Levine, Emma Brunskill, and Zoran Popovic.
\newblock Offline policy evaluation across representations with applications to
  educational games.
\newblock In \emph{AAMAS}, pages 1077--1084, 2014.

\bibitem[Murphy et~al.(2001)Murphy, van~der Laan, Robins, and
  Group]{murphy2001marginal}
Susan~A Murphy, Mark~J van~der Laan, James~M Robins, and Conduct Problems
  Prevention~Research Group.
\newblock Marginal mean models for dynamic regimes.
\newblock \emph{Journal of the American Statistical Association}, 96\penalty0
  (456):\penalty0 1410--1423, 2001.

\bibitem[Mnih et~al.(2015)Mnih, Kavukcuoglu, Silver, Rusu, Veness, Bellemare,
  Graves, Riedmiller, Fidjeland, Ostrovski, et~al.]{mnih2015human}
Volodymyr Mnih, Koray Kavukcuoglu, David Silver, Andrei~A Rusu, Joel Veness,
  Marc~G Bellemare, Alex Graves, Martin Riedmiller, Andreas~K Fidjeland, Georg
  Ostrovski, et~al.
\newblock Human-level control through deep reinforcement learning.
\newblock \emph{nature}, 518\penalty0 (7540):\penalty0 529--533, 2015.

\bibitem[Lillicrap et~al.(2015)Lillicrap, Hunt, Pritzel, Heess, Erez, Tassa,
  Silver, and Wierstra]{lillicrap2015continuous}
Timothy~P Lillicrap, Jonathan~J Hunt, Alexander Pritzel, Nicolas Heess, Tom
  Erez, Yuval Tassa, David Silver, and Daan Wierstra.
\newblock Continuous control with deep reinforcement learning.
\newblock \emph{arXiv preprint arXiv:1509.02971}, 2015.

\bibitem[Espeholt et~al.(2018)Espeholt, Soyer, Munos, Simonyan, Mnih, Ward,
  Doron, Firoiu, Harley, Dunning, Legg, and Kavukcuoglu]{pmlr-v80-espeholt18a}
Lasse Espeholt, Hubert Soyer, Remi Munos, Karen Simonyan, Vlad Mnih, Tom Ward,
  Yotam Doron, Vlad Firoiu, Tim Harley, Iain Dunning, Shane Legg, and Koray
  Kavukcuoglu.
\newblock {IMPALA}: Scalable distributed deep-{RL} with importance weighted
  actor-learner architectures.
\newblock In Jennifer Dy and Andreas Krause, editors, \emph{Proceedings of the
  35th International Conference on Machine Learning}, volume~80 of
  \emph{Proceedings of Machine Learning Research}, pages 1407--1416. PMLR,
  10--15 Jul 2018.
\newblock URL \url{http://proceedings.mlr.press/v80/espeholt18a.html}.

\bibitem[Lange et~al.(2012)Lange, Gabel, and Riedmiller]{lange2012batch}
Sascha Lange, Thomas Gabel, and Martin Riedmiller.
\newblock Batch reinforcement learning.
\newblock In \emph{Reinforcement learning}, pages 45--73. Springer, 2012.

\bibitem[Jaques et~al.(2019)Jaques, Ghandeharioun, Shen, Ferguson, Lapedriza,
  Jones, Gu, and Picard]{jaques2019way}
Natasha Jaques, Asma Ghandeharioun, Judy~Hanwen Shen, Craig Ferguson, Agata
  Lapedriza, Noah Jones, Shixiang Gu, and Rosalind Picard.
\newblock Way off-policy batch deep reinforcement learning of implicit human
  preferences in dialog.
\newblock \emph{arXiv preprint arXiv:1907.00456}, 2019.

\bibitem[Fujimoto et~al.(2019)Fujimoto, Meger, and Precup]{fujimoto2019off}
Scott Fujimoto, David Meger, and Doina Precup.
\newblock Off-policy deep reinforcement learning without exploration.
\newblock In \emph{International Conference on Machine Learning}, pages
  2052--2062. PMLR, 2019.

\bibitem[Bertsekas and Tsitsiklis(1996)]{bertsekas_neuro}
Dimitri~P. Bertsekas and John~N. Tsitsiklis.
\newblock \emph{Neuro-Dynamic Programming}.
\newblock Athena Scientific, 1st edition, 1996.
\newblock ISBN 1886529108.

\bibitem[Kushner and Yin(2003)]{kushner2003}
Harold Kushner and George Yin.
\newblock \emph{Stochastic Approximation and Recursive Algorithms and
  Applications}.
\newblock Springer-Verlag, 2003.
\newblock \doi{10.1007/b97441}.

\bibitem[Fran{\c{c}}ois-Lavet et~al.(2019)Fran{\c{c}}ois-Lavet, Rabusseau,
  Pineau, Ernst, and Fonteneau]{francois2019overfitting}
Vincent Fran{\c{c}}ois-Lavet, Guillaume Rabusseau, Joelle Pineau, Damien Ernst,
  and Raphael Fonteneau.
\newblock On overfitting and asymptotic bias in batch reinforcement learning
  with partial observability.
\newblock \emph{Journal of Artificial Intelligence Research}, 65:\penalty0
  1--30, 2019.

\bibitem[Sinha and Garg(2021)]{sinha2021s4rl}
Samarth Sinha and Animesh Garg.
\newblock S4rl: Surprisingly simple self-supervision for offline reinforcement
  learning.
\newblock \emph{arXiv preprint arXiv:2103.06326}, 2021.

\bibitem[Agarwal et~al.(2020)Agarwal, Schuurmans, and
  Norouzi]{agarwal2020optimistic}
Rishabh Agarwal, Dale Schuurmans, and Mohammad Norouzi.
\newblock An optimistic perspective on offline reinforcement learning.
\newblock In \emph{International Conference on Machine Learning}, pages
  104--114. PMLR, 2020.

\bibitem[Rebba et~al.(2006)Rebba, Mahadevan, and Huang]{rebba2006validation}
Ramesh Rebba, Sankaran Mahadevan, and Shuping Huang.
\newblock Validation and error estimation of computational models.
\newblock \emph{Reliability Engineering \&amp; System Safety}, 91\penalty0
  (10-11):\penalty0 1390--1397, 2006.

\bibitem[Yu et~al.(2019)Yu, Liu, and Nemati]{yu2019reinforcement}
Chao Yu, Jiming Liu, and Shamim Nemati.
\newblock Reinforcement learning in healthcare: a survey.
\newblock \emph{arXiv preprint arXiv:1908.08796}, 2019.

\bibitem[Gottesman et~al.(2019)Gottesman, Johansson, Komorowski, Faisal,
  Sontag, Doshi-Velez, and Celi]{Gottesman2019}
Omer Gottesman, Fredrik Johansson, Matthieu Komorowski, Aldo Faisal, David
  Sontag, Finale Doshi-Velez, and Leo~Anthony Celi.
\newblock Guidelines for reinforcement learning in healthcare.
\newblock \emph{Nature Medicine}, 25\penalty0 (1):\penalty0 16--18, January
  2019.
\newblock \doi{10.1038/s41591-018-0310-5}.
\newblock URL \url{https://doi.org/10.1038/s41591-018-0310-5}.

\bibitem[Kumar et~al.(2019)Kumar, Fu, Soh, Tucker, and
  Levine]{kumar2019stabilizing}
Aviral Kumar, Justin Fu, Matthew Soh, George Tucker, and Sergey Levine.
\newblock Stabilizing off-policy q-learning via bootstrapping error reduction.
\newblock In \emph{Advances in Neural Information Processing Systems}, pages
  11784--11794, 2019.

\bibitem[Wu et~al.(2019)Wu, Tucker, and Nachum]{wu2019behavior}
Yifan Wu, George Tucker, and Ofir Nachum.
\newblock Behavior regularized offline reinforcement learning.
\newblock \emph{arXiv preprint arXiv:1911.11361}, 2019.

\bibitem[Wang et~al.(2020)Wang, Novikov, {\.Z}o{\l}na, Springenberg, Reed,
  Shahriari, Siegel, Merel, Gulcehre, Heess, et~al.]{wang2020critic}
Ziyu Wang, Alexander Novikov, Konrad {\.Z}o{\l}na, Jost~Tobias Springenberg,
  Scott Reed, Bobak Shahriari, Noah Siegel, Josh Merel, Caglar Gulcehre,
  Nicolas Heess, et~al.
\newblock Critic regularized regression.
\newblock In \emph{Advances in Neural Information Processing Systems}, 2020.

\bibitem[Johnson et~al.(2016)Johnson, Pollard, Shen, Li-wei, Feng, Ghassemi,
  Moody, Szolovits, Celi, and Mark]{JohnsonPollardShenEtAl2016}
Alistair~EW Johnson, Tom~J Pollard, Lu~Shen, H~Lehman Li-wei, Mengling Feng,
  Mohammad Ghassemi, Benjamin Moody, Peter Szolovits, Leo~Anthony Celi, and
  Roger~G Mark.
\newblock {MIMIC-III}, a freely accessible critical care database.
\newblock \emph{Scientific data}, 3:\penalty0 160035, 2016.

\bibitem[Henry et~al.(2015)Henry, Hager, Pronovost, and
  Saria]{henry2015targeted}
Katharine~E Henry, David~N Hager, Peter~J Pronovost, and Suchi Saria.
\newblock A targeted real-time early warning score (trewscore) for septic
  shock.
\newblock \emph{Science translational medicine}, 7\penalty0 (299):\penalty0
  299ra122--299ra122, 2015.

\bibitem[Futoma et~al.(2017)Futoma, Hariharan, Heller, Sendak, Brajer, Clement,
  Bedoya, and O’Brien]{futoma2017improved}
Joseph Futoma, Sanjay Hariharan, Katherine Heller, Mark Sendak, Nathan Brajer,
  Meredith Clement, Armando Bedoya, and Cara O’Brien.
\newblock An improved multi-output gaussian process rnn with real-time
  validation for early sepsis detection.
\newblock In \emph{Machine Learning for Healthcare Conference}, pages 243--254,
  2017.

\bibitem[Komorowski et~al.(2018)Komorowski, Celi, Badawi, Gordon, and
  Faisal]{komorowski2018artificial}
Matthieu Komorowski, Leo~A Celi, Omar Badawi, Anthony~C Gordon, and A~Aldo
  Faisal.
\newblock The artificial intelligence clinician learns optimal treatment
  strategies for sepsis in intensive care.
\newblock \emph{Nature medicine}, 24\penalty0 (11):\penalty0 1716--1720, 2018.

\bibitem[Saria(2018)]{saria2018individualized}
Suchi Saria.
\newblock Individualized sepsis treatment using reinforcement learning.
\newblock \emph{Nature medicine}, 24\penalty0 (11):\penalty0 1641--1642, 2018.

\bibitem[Deutschman and Tracey(2014)]{DEUTSCHMAN2014463}
Clifford S. Deutschman and Kevin J. Tracey.
\newblock Sepsis: Current dogma and new perspectives.
\newblock \emph{Immunity}, 40\penalty0 (4):\penalty0 463 -- 475, 2014.
\newblock ISSN 1074-7613.
\newblock \doi{https://doi.org/10.1016/j.immuni.2014.04.001}.
\newblock URL
  \url{http://www.sciencedirect.com/science/article/pii/S1074761314001150}.

\bibitem[Singer et~al.(2016)Singer, Deutschman, Seymour, Shankar-Hari, Annane,
  Bauer, Bellomo, Bernard, Chiche, Coopersmith, Hotchkiss, Levy, Marshall,
  Martin, Opal, Rubenfeld, van~der Poll, Vincent, and Angus]{Singer_2016}
Mervyn Singer, Clifford~S. Deutschman, Christopher~Warren Seymour, Manu
  Shankar-Hari, Djillali Annane, Michael Bauer, Rinaldo Bellomo, Gordon~R.
  Bernard, Jean-Daniel Chiche, Craig~M. Coopersmith, Richard~S. Hotchkiss,
  Mitchell~M. Levy, John~C. Marshall, Greg~S. Martin, Steven~M. Opal, Gordon~D.
  Rubenfeld, Tom van~der Poll, Jean-Louis Vincent, and Derek~C. Angus.
\newblock The third international consensus definitions for sepsis and septic
  shock (sepsis-3).
\newblock \emph{{JAMA}}, 315\penalty0 (8):\penalty0 801, feb 2016.
\newblock \doi{10.1001/jama.2016.0287}.
\newblock URL \url{https://doi.org/10.1001%2Fjama.2016.0287}.

\bibitem[Vincent et~al.(2013)Vincent, Opal, Marshall, and
  Tracey]{VINCENT2013774}
Jean-Louis Vincent, Steven~M Opal, John~C Marshall, and Kevin~J Tracey.
\newblock Sepsis definitions: time for change.
\newblock \emph{The Lancet}, 381\penalty0 (9868):\penalty0 774 -- 775, 2013.
\newblock ISSN 0140-6736.
\newblock \doi{https://doi.org/10.1016/S0140-6736(12)61815-7}.
\newblock URL
  \url{http://www.sciencedirect.com/science/article/pii/S0140673612618157}.

\bibitem[Torio and Moore(2016)]{hcup:2013}
C~Torio and B~Moore.
\newblock National inpatient hospital costs: The most expensive conditions by
  payer, 2013. {Statistical Brief} \#204.
\newblock \emph{Healthcare Cost and Utilization Project (HCUP) Statistical
  Briefs}, May 2016.
\newblock URL
  \url{http://www.hcup-us.ahrq.gov/reports/statbriefs/sb204-Most-Expensive-Hospital-Conditions.pdf}.

\bibitem[Martin(2012)]{martin2012sepsis}
Greg~S Martin.
\newblock Sepsis, severe sepsis and septic shock: changes in incidence,
  pathogens and outcomes.
\newblock \emph{Expert review of anti-infective therapy}, 10\penalty0
  (6):\penalty0 701--706, 2012.

\bibitem[Marik et~al.(2017)Marik, Linde-Zwirble, Bittner, Sahatjian, and
  Hansell]{pmid28130687}
P.~E. Marik, W.~T. Linde-Zwirble, E.~A. Bittner, J.~Sahatjian, and D.~Hansell.
\newblock {{F}luid administration in severe sepsis and septic shock, patterns
  and outcomes: an analysis of a large national database}.
\newblock \emph{Intensive Care Med}, 43\penalty0 (5):\penalty0 625--632, May
  2017.

\bibitem[Waechter et~al.(2014)Waechter, Kumar, Lapinsky, Marshall, Dodek,
  Arabi, Parrillo, Dellinger, Garland, Dial, Ellis, Feinstein, Gurka, Guzman,
  Keenan, Kramer, Kumar, Laporta, Laupland, Light, Maki, Martinka, Memish,
  Mirzanejad, Patel, Penner, Roberts, Ronald, Simon, Sharma, Shirawi, Skrobik,
  Wood, Wood, Zanotti, Ahsan, Bahrainian, Bohmeier, Carter, Chou, Delgra,
  Egbujuo, Fu, Gonzales, Gulati, Halmarson, Hansen, Haque, Harvey, Khan,
  Kolesar, Kravetsky, Kumar, Merali, Muggaberg, Paulin, Peters, Richards,
  Schorr, Serrano, Suleman, Singh, Sullivan, Suppes, Taiberg, Tchokonte,
  Torshizi, and Wiebe]{pmid25072761}
J.~Waechter, A.~Kumar, S.~E. Lapinsky, J.~Marshall, P.~Dodek, Y.~Arabi, J.~E.
  Parrillo, R.~P. Dellinger, A.~Garland, S.~Dial, P.~Ellis, D.~Feinstein,
  D.~Gurka, J.~Guzman, S.~Keenan, A.~Kramer, A.~Kumar, D.~Laporta, K.~Laupland,
  B.~Light, D.~Maki, G.~Martinka, Z.~Memish, Y.~Mirzanejad, G.~Patel,
  C.~Penner, D.~Roberts, J.~Ronald, D.~Simon, S.~Sharma, N.~A. Shirawi,
  Y.~Skrobik, G.~Wood, K.~E. Wood, S.~Zanotti, M.~W. Ahsan, M.~Bahrainian,
  R.~Bohmeier, L.~Carter, H.~Chou, S.~Delgra, C.~Egbujuo, W.~Fu, C.~Gonzales,
  H.~Gulati, E.~Halmarson, J.~Hansen, Z.~Haque, J.~Harvey, F.~Khan, L.~Kolesar,
  L.~Kravetsky, R.~Kumar, N.~Merali, S.~Muggaberg, H.~Paulin, C.~Peters,
  J.~Richards, C.~Schorr, H.~Serrano, M.~Suleman, A.~Singh, K.~Sullivan,
  R.~Suppes, L.~Taiberg, R.~Tchokonte, O.~A. Torshizi, and K.~Wiebe.
\newblock {{I}nteraction between fluids and vasoactive agents on mortality in
  septic shock: a multicenter, observational study}.
\newblock \emph{Crit. Care Med.}, 42\penalty0 (10):\penalty0 2158--2168, Oct
  2014.

\bibitem[Gauer et~al.(2020)Gauer, Forbes, and Boyer]{gauer2020sepsis}
Robert Gauer, Damon Forbes, and Nathan Boyer.
\newblock Sepsis: diagnosis and management.
\newblock \emph{American family physician}, 101\penalty0 (7):\penalty0
  409--418, 2020.

\bibitem[Raghu et~al.(2017)Raghu, Komorowski, Celi, Szolovits, and
  Ghassemi]{raghu2017continuous}
Aniruddh Raghu, Matthieu Komorowski, Leo~Anthony Celi, Peter Szolovits, and
  Marzyeh Ghassemi.
\newblock Continuous state-space models for optimal sepsis treatment: a deep
  reinforcement learning approach.
\newblock In \emph{Machine Learning for Healthcare Conference}, pages 147--163.
  PMLR, 2017.

\bibitem[Peng et~al.(2018)Peng, Ding, Wihl, Gottesman, Komorowski, Lehman,
  Ross, Faisal, and Doshi-Velez]{peng2018improving}
Xuefeng Peng, Yi~Ding, David Wihl, Omer Gottesman, Matthieu Komorowski,
  Li-wei~H Lehman, Andrew Ross, Aldo Faisal, and Finale Doshi-Velez.
\newblock Improving sepsis treatment strategies by combining deep and
  kernel-based reinforcement learning.
\newblock In \emph{AMIA Annual Symposium Proceedings}, volume 2018, page 887.
  American Medical Informatics Association, 2018.

\bibitem[Li et~al.(2019)Li, Komorowski, and Faisal]{li2019optimizing}
Luchen Li, Matthieu Komorowski, and Aldo~A Faisal.
\newblock Optimizing sequential medical treatments with auto-encoding heuristic
  search in {POMDP}s.
\newblock \emph{arXiv preprint arXiv:1905.07465}, 2019.

\bibitem[Tang et~al.(2020)Tang, Modi, Sjoding, and Wiens]{tang2020clinician}
Shengpu Tang, Aditya Modi, Michael Sjoding, and Jenna Wiens.
\newblock Clinician-in-the-loop decision making: Reinforcement learning with
  near-optimal set-valued policies.
\newblock In \emph{International Conference on Machine Learning}, pages
  9387--9396. PMLR, 2020.

\bibitem[Garc{\i}a and Fern{\'a}ndez(2015)]{garcia2015comprehensive}
Javier Garc{\i}a and Fernando Fern{\'a}ndez.
\newblock A comprehensive survey on safe reinforcement learning.
\newblock \emph{Journal of Machine Learning Research}, 16\penalty0
  (1):\penalty0 1437--1480, 2015.

\bibitem[Thomas(2015)]{thomas2015safe}
Philip~S Thomas.
\newblock \emph{Safe reinforcement learning}.
\newblock PhD thesis, University of Massachusetts Libraries, 2015.

\bibitem[Hadfield-Menell et~al.(2016)Hadfield-Menell, Dragan, Abbeel, and
  Russell]{hadfield2016cooperative}
Dylan Hadfield-Menell, Anca Dragan, Pieter Abbeel, and Stuart Russell.
\newblock Cooperative inverse reinforcement learning.
\newblock In \emph{Proceedings of the 30th International Conference on Neural
  Information Processing Systems}, pages 3916--3924, 2016.

\bibitem[Hadfield-Menell et~al.(2017)Hadfield-Menell, Milli, Abbeel, Russell,
  and Dragan]{hadfield2017inverse}
Dylan Hadfield-Menell, Smitha Milli, Pieter Abbeel, Stuart Russell, and Anca~D
  Dragan.
\newblock Inverse reward design.
\newblock In \emph{Proceedings of the 31st International Conference on Neural
  Information Processing Systems}, pages 6768--6777, 2017.

\bibitem[Thomas et~al.(2019)Thomas, da~Silva, Barto, Giguere, Brun, and
  Brunskill]{thomas2019preventing}
Philip~S Thomas, Bruno~Castro da~Silva, Andrew~G Barto, Stephen Giguere, Yuriy
  Brun, and Emma Brunskill.
\newblock Preventing undesirable behavior of intelligent machines.
\newblock \emph{Science}, 366\penalty0 (6468):\penalty0 999--1004, 2019.

\bibitem[Laroche et~al.(2019)Laroche, Trichelair, and Tachet~des
  Combes]{laroche2019safe}
Romain Laroche, Paul Trichelair, and Remi Tachet~des Combes.
\newblock Safe policy improvement with baseline bootstrapping.
\newblock In \emph{International Conference on Machine Learning (ICML)}, June
  2019.

\bibitem[Fatemi et~al.(2019)Fatemi, Sharma, Van~Seijen, and
  Kahou]{fatemi2019dead}
Mehdi Fatemi, Shikhar Sharma, Harm Van~Seijen, and Samira~Ebrahimi Kahou.
\newblock Dead-ends and secure exploration in reinforcement learning.
\newblock In \emph{International Conference on Machine Learning}, pages
  1873--1881, 2019.

\bibitem[Irpan et~al.(2019)Irpan, Rao, Bousmalis, Harris, Ibarz, and
  Levine]{irpan2019off}
Alexander Irpan, Kanishka Rao, Konstantinos Bousmalis, Chris Harris, Julian
  Ibarz, and Sergey Levine.
\newblock Off-policy evaluation via off-policy classification.
\newblock \emph{Advances in Neural Information Processing Systems},
  32:\penalty0 5437--5448, 2019.

\bibitem[Maley et~al.(2020)Maley, Wanis, Young, and Celi]{maley2020mortality}
Jason~H Maley, Kerollos~N Wanis, Jessica~G Young, and Leo~A Celi.
\newblock Mortality prediction models, causal effects, and end-of-life decision
  making in the intensive care unit.
\newblock \emph{BMJ Health \&amp; Care Informatics}, 27\penalty0 (3), 2020.

\bibitem[Johnson et~al.(2018)Johnson, Stone, Celi, and Pollard]{mimicweb}
Alistair~Ew Johnson, David~J Stone, Leo~A Celi, and Tom~J Pollard.
\newblock The {MIMIC} code repository: enabling reproducibility in critical
  care research.
\newblock \emph{J. Am. Med. Inform. Assoc.}, 25\penalty0 (1):\penalty0 32--39,
  January 2018.

\bibitem[Killian et~al.(2020)Killian, Zhang, Subramanian, Fatemi, and
  Ghassemi]{killian2020empirical}
Taylor~W Killian, Haoran Zhang, Jayakumar Subramanian, Mehdi Fatemi, and
  Marzyeh Ghassemi.
\newblock An empirical study of representation learning for reinforcement
  learning in healthcare.
\newblock In \emph{Machine Learning for Health}, pages 139--160. PMLR, 2020.
\newblock URL \url{http://proceedings.mlr.press/v136/killian20a}.

\bibitem[Subramanian and Mahajan(2019)]{subramanian2019approximate}
Jayakumar Subramanian and Aditya Mahajan.
\newblock Approximate information state for partially observed systems.
\newblock In \emph{Conference of Decision and Control (CDC), Nice, France},
  2019.

\bibitem[Hasselt et~al.(2016)Hasselt, Guez, and Silver]{Hasselt2016}
Hado~van Hasselt, Arthur Guez, and David Silver.
\newblock Deep reinforcement learning with double q-learning.
\newblock In \emph{Proceedings of the Thirtieth AAAI Conference on Artificial
  Intelligence}, AAAI’16, page 2094–2100. AAAI Press, 2016.

\bibitem[Maaten and Hinton(2008)]{maaten2008visualizing}
Laurens van~der Maaten and Geoffrey Hinton.
\newblock Visualizing data using t-sne.
\newblock \emph{Journal of machine learning research}, 9\penalty0
  (Nov):\penalty0 2579--2605, 2008.

\bibitem[Vincent et~al.(2014)Vincent, Marshall, Ñamendys Silva, François,
  Martin-Loeches, Lipman, Reinhart, Antonelli, Pickkers, Njimi, Jimenez, and
  Sakr]{VINCENT2014380}
Jean-Louis Vincent, John~C Marshall, Silvio~A Ñamendys Silva, Bruno François,
  Ignacio Martin-Loeches, Jeffrey Lipman, Konrad Reinhart, Massimo Antonelli,
  Peter Pickkers, Hassane Njimi, Edgar Jimenez, and Yasser Sakr.
\newblock Assessment of the worldwide burden of critical illness: the intensive
  care over nations (icon) audit.
\newblock \emph{The Lancet Respiratory Medicine}, 2\penalty0 (5):\penalty0 380
  -- 386, 2014.
\newblock ISSN 2213-2600.
\newblock \doi{https://doi.org/10.1016/S2213-2600(14)70061-X}.
\newblock URL
  \url{http://www.sciencedirect.com/science/article/pii/S221326001470061X}.

\bibitem[Fleischmann et~al.(2016)Fleischmann, Scherag, Adhikari, Hartog,
  Tsaganos, Schlattmann, Angus, and Reinhart]{pmid26414292}
C.~Fleischmann, A.~Scherag, N.~K. Adhikari, C.~S. Hartog, T.~Tsaganos,
  P.~Schlattmann, D.~C. Angus, and K.~Reinhart.
\newblock {{A}ssessment of {G}lobal {I}ncidence and {M}ortality of
  {H}ospital-treated {S}epsis. {C}urrent {E}stimates and {L}imitations}.
\newblock \emph{Am. J. Respir. Crit. Care Med.}, 193\penalty0 (3):\penalty0
  259--272, Feb 2016.

\bibitem[Levine et~al.(2020)Levine, Kumar, Tucker, and Fu]{levine2020offline}
Sergey Levine, Aviral Kumar, George Tucker, and Justin Fu.
\newblock Offline reinforcement learning: Tutorial, review, and perspectives on
  open problems.
\newblock \emph{arXiv preprint arXiv:2005.01643}, 2020.

\bibitem[Carroll(2017)]{carroll2017high}
Aaron~E Carroll.
\newblock The high costs of unnecessary care.
\newblock \emph{Jama}, 318\penalty0 (18):\penalty0 1748--1749, 2017.

\bibitem[Cho et~al.(2014)Cho, Van~Merri{\"e}nboer, Bahdanau, and
  Bengio]{cho2014properties}
Kyunghyun Cho, Bart Van~Merri{\"e}nboer, Dzmitry Bahdanau, and Yoshua Bengio.
\newblock On the properties of neural machine translation: Encoder-decoder
  approaches.
\newblock \emph{arXiv preprint arXiv:1409.1259}, 2014.

\bibitem[Komorowski(2018)]{aiclinician}
Matthieu Komorowski.
\newblock {AI C}linician.
\newblock \url{https://github.com/matthieukomorowski/AI_Clinician}, 2018.
\newblock Accessed: 2019-08-16.

\end{thebibliography}
\bibliographystyle{unsrtnat}

\clearpage

\appendix

\setcounter{page}{1}

\renewcommand{\baselinestretch}{1} 

\renewcommand\thesection{A\arabic{section}}

\renewcommand{\figurename}{}
\renewcommand\thefigure{Fig. A\arabic{figure}}
\setcounter{figure}{0}

\renewcommand{\tablename}{}
\renewcommand\thetable{Table A\arabic{table}}
\setcounter{table}{0}

\setcounter{section}{0}

\newgeometry{left=1in,bottom=1in}

\section{Formal Results and Proofs} \label{appendix:sec:proofs}

For simplicity, in all the arguments below, we refer to a positive terminal state as \textit{recovery} and a negative terminal state as \textit{death}. As with the main text, we also use \textit{treatment} in place of \textit{action}, which is the common term in RL texts. The rest of terminology follows the definitions presented in the main text. 

\raggedright\textbf{Lemma 1.}\setlength{\parindent}{15pt}
\begin{enumerate}[L{1}.1.]
    \item $V_{D}^{*}(s)=Q_{D}^{*}(s,a)=-1$ for all the treatments $a$ if and only if $s$ is a dead-end.
    \item $V_{R}^{*}(s)=\max_{a}Q_{R}^{*}(s,a)=1$ if and only if $s$ is a rescue.
\end{enumerate} 

\raggedright\textbf{Proof.}\setlength{\parindent}{15pt} To prove the first part of the lemma, we assume $s$ is a dead-end and prove $Q_{D}^{*}(s,a)=-1$ for all the treatments. The definition of return directly implies that the return of all the trajectories from $s$ is precisely $-1$ since all of them reach a death terminal state and $\gamma=1$. The expected return is therefore also $-1$ regardless of stochasticity; hence, $Q_{D}^{*}(s,a)=-1$. 

Conversely, let for a given state $s$ we have $Q_{D}^{*}(s,a)=-1$ for all treatments $a$. We next prove that $s$ is a dead-end. For a transition $(s,a,s')$, the next state $s'$ is either of a non-terminal state with $r_{D}(s,a,s')=0$, $max_{a'}Q_{D}^{*}(s',a')=-1$; a non-terminal state with $r_{D}(s,a,s')=0$, $max_{a'}Q_{D}^{*}(s',a')> -1$; a death terminal state (i.e., $r_{D}(s,a,s')=-1$, $Q_{D}^{*}(s',a')=0~~\forall a'$); or a recovery terminal state (i.e., $r_{D}(s,a,s')=0$, $Q_{D}^{*}(s',a')=0~~\forall a'$). 

Let $C_{R}$ and $C_{N}$ denote respectively the sets of ``recovery terminal states'' and ``non-terminal states $s'$ with $max_{a'}Q_{D}^{*}(s',a')>-1$''. Note that $C_{R}$ and $C_{N}$ are disjoint, and that if a state $s'$ is not in $C_{R}\cup C_{N}$ then it is either a death terminal state (hence $r_{D}(\cdot, \cdot, s')=-1$ and $Q^{*}_{D}(s',\cdot)=0$), or a non-terminal with -1 value (hence $r_{D}(\cdot, \cdot, s')=0$ and $Q^{*}_{D}(s',\cdot)=-1$). Using Bellman equation, we write

\begin{align}
    \nonumber -1 = Q_{D}^{*}(s,a)&=\sum_{s'}T(s,a,s')[r_{D}(s,a,s') + \max_{a'}Q_{D}^{*}(s',a')] \\
    \nonumber &=\sum_{s'\notin C_{R}\cup C_{N}}{T(s,a,s')\times -1} + \sum_{s'\in C_{R}}{T(s,a,s')\times 0} + \sum_{s'\in C_{N}}{T(s,a,s') \max_{a'}Q_{D}^{*}(s', a')} \\
    \nonumber &=-\left[1-\sum_{s'\in C_{R}\cup C_{N}}{T(s,a,s')}\right] + \sum_{s'\in C_{N}}{T(s,a,s') \max_{a'}Q_{D}^{*}(s', a')} \\
    &=-1 + \sum_{s'\in C_{R}}{T(s,a,s')} + \sum_{s'\in C_{N}}{T(s,a,s') \left[ 1+\max_{a'}Q_{D}^{*}(s', a')\right]} \label{eq:proof_lemma1} 
\end{align}
Because $T(s,a,s')$ is non-negative it therefore must be zero for both all $s'\in C_{R}$ and all $s'\in C_{N}$ (in the last line $\max_{a'}Q_{D}^{*}(s', a')\neq -1$). Hence, the next state is either a death terminal state or a non-terminal state with $Q_{D}^{*}(s', \cdot)$ of precisely $-1$ for all the treatments. Continuing with the same line of argument, it therefore follows that if $Q_{D}^{*}(s,a)=-1$ then all possible trajectories after $(s,a)$ will reach a death terminal state and all states on such trajectories assume the value of -1. Finally, if $V_{D}^{*}(s)=-1$ then $max_{a}Q_{D}^{*}(s,a)=-1$, which implies $Q_{D}^{*}(s,a)=-1$ for all $a$. It therefore follows that all trajectories from $s$ will reach a death terminal state, and by definition $s$ is a dead-end.

To prove L1.2, for the sufficiency we cannot use a similar argument as for L1.1 since not all the returns are $+1$; only the maximum needs to be $+1$. If $s$ is a rescue state, then by definition there must exist at least one trajectory w.p.1 to recovery. Starting from the last state before recovery on such a trajectory, we go backward and invoke Bellman equation. For the last state-treatment $(s'',a'')$ that transitions to recovery we have $Q_{R}^{*}(s'',a'')=+1$, hence $max_{a'}Q_{R}^{*}(s'',a')=+1$. Similarly, for all other states $s'$ on the deterministic trajectory to recovery, we conclude that $max_{a'}Q_{R}^{*}(s',a')=+1$, which implies $max_{a'}Q_{R}^{*}(s,a')=+1$, as stated in the lemma. 

For the necessity, the argument is similar to that of L1.1. In particular, we can show in a similar way as in L1.1 that if $Q_{R}^{*}(s,a)=1$ then $\max_{a'} Q_{R}^{*}(s',a')=1$ for all the immediate next states $s'$ after $(s,a)$ if they are non-terminal. It implies that if $s'$ is non-terminal, then at least one treatment exists whose value at $s'$ is $+1$. Furthermore, for all state-treatment pairs whose values are $+1$, if the treatment causes transitioning to a terminal state it deterministically must be recovery (i.e., it cannot be recovery w.p. $p$ and death w.p. $1-p$). We therefore conclude that there is at least one trajectory from $s$ to recovery with probability one; hence $s$ is a rescue state.  
\hfill $\blacksquare$

\raggedright\textbf{Lemma 2.}\setlength{\parindent}{15pt} Let treatment $a$ be administered at state $s$, and $F_{D}(s,a)$ and $F_{R}(s,a)$ denote the probability that the next state will be terminal with death or recovery, respectively. Let further $P_{D}(s, a)$ and $P_{R}(s, a)$ denote the probability of transitioning to a dead-end or a rescue state, respectively,  i.e. $P_{D}(s, a) = \sum_{s'\in\mathcal{S}_{D}}T(s, a, s')$ and $P_{R}(s,a) = \sum_{s'\in\mathcal{S}_{R}}T(s, a, s')$. Let $M_{D}(s, a)$ be the probability that the next state is neither a dead-end nor immediate death, and the patient ultimately expires while all the treatments are selected according to the greedy policy with respect to $Q^{*}_{D}$. Similarly, let $M_{R}(s, a)$ be the probability that the next state is neither immediate recovery nor a rescue state, but the patient ultimately recovers while all future treatments are selected according to the greedy policy with respect to $Q^{*}_{R}$. We have
\begin{enumerate}[L{2}.1.]
\item $-Q_{D}^{*}(s,a) = P_{D}(s,a) + M_{D}(s,a) + F_{D}(s,a)$
\item $Q_{R}^{*}(s,a) = P_{R}(s,a) + M_{R}(s,a) + F_{R}(s,a)$
\end{enumerate}

\raggedright\textbf{Proof.}\setlength{\parindent}{15pt} For the first part, Bellman equation reads as the following:
\begin{equation}
    Q_{D}^{*}(s,a)=\sum_{s'}T(s,a,s')[r_{D}(s,a,s') + \max_{a'}Q_{D}^{*}(s',a')]
\end{equation}

The next state $s'$ is either of the following:
\begin{enumerate}
    \item a dead-end state, where $r_{D}(s,a,s')=0$; $Q_{D}(s',a')=-1, ~~\forall a'$ (due to Lemma 1); and $\sum_{s'}T(s,a,s') = P_{D}(s,a)$,
    \item a death terminal state, where $r_{D}(s,a,s')=-1$; $Q_{D}(s',a')=0, ~~\forall a'$; and $\sum_{s'}T(s,a,s') = F_{D}(s,a)$,
    \item a recovery terminal state where $r_{D}(s,a,s')=0$, and $Q_{D}(s',a')=0, ~~\forall a'$, and
    \item a non-terminal, non dead-end state, where  $r_{D}(s,a,s')=0$.
\end{enumerate}

Item 3 vanishes and items 1 and 2 result in the first and the last terms in L2.1. For the item 4 above, assume any treatment $a'$ at the state $s'$ and consider all the possible roll-outs starting from $(s', a')$ under execution of the greedy policy w.r.t. $Q_{D}^{*}$ (which maximally avoids future mortality). At the end of each roll-out, the roll-out trajectory necessarily either reaches death with the $\mathcal{M}_{D}$ return of $-1$ for the trajectory, or it reaches recovery with the $\mathcal{M}_{D}$ return of $0$ for the trajectory. Hence, the expected return from $(s', a')$ will be $-1$ times the sum of probabilities of all the roll-outs that reach death (plus zero times sum of the rest). That is, $Q_{D}^{*}(s',a')$ is \emph{the negative total probability of future death} from $(s',a')$ if optimal treatments (w.r.t. $Q_{D}^{*}$) are always known and administered afterwards. Consequently, $\max_{a'} Q_{D}^{*}(s',a')$ would be \emph{negative minimum probability of future death} from state $s'$, again if optimal treatments are known and administered at $s'$ and afterwards. Therefore, $\sum_{s'}T(s,a,s') \max_{a'} Q_{D}^{*}(s',a')$ is negative minimum probability of future death from $(s,a)$ under optimal policy, which by definition is $-M_{D}(s,a)$. This shapes the middle term of L2.1 and concludes the proof.

The second part follows a similar argument. In particular, $Q_{R}^{*}(s',a')$ is the probability of reaching recovery under the execution of greedy policy w.r.t. $Q_{R}^{*}$ (which itself maximizes reaching a recovery terminal). Therefore, $\max_{a'}Q_{R}^{*}(s',a')$ is the maximum probability of reaching recovery under optimal policy from $s'$, and finally $\sum_{s'}T(s,a,s') \max_{a'} Q_{R}^{*}(s',a')$ induces maximum probability of reaching recovery from $(s,a)$.

\hfill $\blacksquare$

\raggedright\textbf{Lemma 3.}\setlength{\parindent}{15pt}

\begin{enumerate}[L{3}.1.]
\item State $s$ is a dead-end if and only if $P_{D}(s,a) + F_{D}(s,a)=1$ for \emph{all} treatments $a$.
\item State $s$ is a rescue if and only if $P_{R}(s,a) + F_{R}(s,a)=1$ for \emph{at least one treatment} $a$.
\end{enumerate}

\raggedright\textbf{Proof.}\setlength{\parindent}{15pt} 
For part one, we note that $P_{D}(s,a)$, $M_{D}(s,a)$, and $F_{D}(s,a)$ are parts of the transition probability to the next state, hence 
$$P_{D}(s,a) + M_{D}(s,a) + F_{D}(s,a)\leq 1$$
Therefore, $P_{D}(s,a) + F_{D}(s,a)=1$ deduces $P_{D}(s,a) + M_{D}(s,a) + F_{D}(s,a)=1$ (i.e., $M_{D}(s,a)=0$). Invoking L2.1 induces $Q^{*}_{D}(s,a)=-1$ for all $a$; hence, $s$ is a dead-end due to L1.1. Conversely, if $s$ is a dead-end, L1.1 induces that $Q^{*}_{D}(s,a)=-1$ for \emph{all treatments} $a$. Invoking \eqref{eq:proof_lemma1} again, it follows that the next state cannot be a recovery terminal state or a non-terminal state with $\max_{a'}Q^{*}_{D}(s',a') > -1$, which implies the next state is either a dead-end or a death terminal state. Hence, $P_{D}(s,a) + F_{D}(s,a)=1$ for all treatments $a$.

Similar proof holds for L3.2. Note that the counterpart of \eqref{eq:proof_lemma1} for this case is as the following:

\begin{align} \label{eq:bellman_rescue}
    1= 1  - \left( \sum_{s'\in C'_{D}}{T(s,a,s')} + \sum_{s'\in C'_{N}}{T(s,a,s') \left[ 1-\max_{a'}Q_{R}^{*}(s', a')\right]}\right)
\end{align}
with $C'_{D}$ and $C'_{N}$ denoting, respectively, the sets of death terminal states and non-terminal states with $\max_{a'}Q^{*}_{R}(s',a')<1$. Similarly, \eqref{eq:bellman_rescue} necessitates $T(s,a,\cdot)$ must be zero for all transitions to $C'_{D}$ and $C'_{N}$. 

\hfill $\blacksquare$

\raggedright\textbf{Theorem 1.}\setlength{\parindent}{15pt} The followings hold:
\begin{enumerate}[T.1]
    \item $P_{D}(s,a)+F_{D}(s,a)=1$ if and only if $Q^{*}_{D}(s,a)=-1$.
    \item $P_{R}(s,a)+F_{R}(s,a)=1$ if and only if $Q^{*}_{R}(s,a)=1$.
    \item There exists a threshold $\delta_{D}\in(-1, 0)$ independent of states and treatments, such that $Q^{*}_{D}(s,a)\ge \delta_{D}$ for all $s$ and $a$, unless if and only if $P_{D}(s,a)+F_{D}(s,a) = 1$.
    \item There exists a threshold $\delta_{R}\in(0, 1)$ independent of states and treatments, such that $Q^{*}_{R}(s,a)\le \delta_{R}$ for all $s$ and $a$, unless if and only if $P_{R}(s,a)+F_{R}(s,a) = 1$.
    \item For any policy $\pi$, state $s$, and treatment $a$, if $\pi(s,a) \le 1+Q^{*}_{D}(s,a)$ and $\lambda\in [0,1]$ exists such that $P_{D}(s,a)+F_{D}(s,a) \ge \lambda$, then $\pi(s,a) \le 1-\lambda$.
    \item For any policy $\pi$, state $s$, and treatment $a$, if $\pi(s,a) \ge Q^{*}_{R}(s,a)$ and $\lambda\in [0,1]$ exists such that $P_{R}(s,a)+F_{R}(s,a) \ge \lambda$, then $\pi(s,a) \ge \lambda$.
\end{enumerate}

\raggedright\textbf{Proof.}\setlength{\parindent}{15pt} (T.1) and (T.2) are immediate from Lemma 1 and 3. For (T.3), it follows from (L1.1) that for a non-dead-end state $s$, we have $Q_{D}^{*}(s, a) > -1$. We choose $\Delta_{D} = \max_{s,a} \left[P_{D}(s,a) + M_{D}(s,a) + F_{D}(s,a)\right]$ for all non-dead-end and non-terminal states $s$ and all treatments $a$. If all the transition probabilities are stationary (or more generically, $\exists \lambda < 1~ : ~ T(s,a,s')< \lambda$ for all non-dead-end and non-terminal transitions) then $\Delta_{D}$ is a fixed value even though it might be very close to $-1$ in principle. As a result, it follows from L2.1 that for any threshold $\delta_{D}\in (-1, -\Delta_{D}]$ we have $Q_{D}^{*}(s, a) \geq -\Delta_{D}$ unless $s$ is a dead-end for which $Q_{D}^{*}(s,a) = -1$ due to L1.1. Furthermore, $\Delta_{D}$ only depends on the transition probabilities $T(s, a, s')$ and not the length of dead-ends. Similar proof concludes (T4).

In order to prove (T.5) and (T.6), we note that both $M_{D}(\cdot, \cdot)$ and $M_{R}(\cdot, \cdot)$ are non-negative for all state-treatments. Using the antecedent of (T.5), $P_{D}(s,a)+F_{D}(s,a)  \ge \lambda$, as well as invoking Lemma 2, it yields:
\begin{align}
    \nonumber Q_{D}^{*}(s,a) &\leq Q_{D}^{*}(s,a) + M_{D}(s,a) \\
    \nonumber &= -\left( P_{D}(s, a) + F_{D}(s,a) \right) \leq -\lambda
\end{align}
which implies $1 + Q_{D}^{*}(s, a) \leq 1-\lambda$. Hence, setting $\pi(s, a) \leq 1 + Q_{D}^{*}(s,a)$ deduces $\pi(s, a) \leq 1-\lambda$.

Similarly, for (T.8) we have $P_{R}(s,a)+F_{R}(s,a) \ge \lambda$, therefore
\begin{align}
    \nonumber Q_{R}^{*}(s,a) &\geq Q_{R}^{*}(s,a) - M_{R}(s,a) \\
    \nonumber &= P_{R}(s, a)+F_{R}(s, a) \geq \lambda
\end{align}
As a result, $\pi(s, a) \geq Q_{R}^{*}(s,a)$ deduces $\pi(s, a) \geq \lambda$.

\hfill $\blacksquare$

\clearpage

\raggedright\textbf{Proposition 1}\setlength{\parindent}{15pt}. Let $Q_{D}(s,a)$ be an approximation of $Q^{*}_{D}(s,a)$, such that
\begin{enumerate}
    \item $Q_{D}(s, a)=Q^{*}_{D}(s,a)=-1$ for all $s\in \mathcal{S}_{D}$.
    \item For all other states, the values satisfy monotonicity with respect to the Bellman operator $\mathcal{T}^{*}$, i.e. $Q_{D}(s,a) \le (\mathcal{T}^{*}Q_{D})(s,a)$ for all $(s,a)$.
    \item All values of $Q_{D}(s,a)$ remain non-positive.
\end{enumerate}  
The security condition still holds if $\pi(s,a) \le 1 + Q_{D}(s,a)$.

\raggedright\textbf{Proof}\setlength{\parindent}{15pt}. 
Using assumptions 1 and 2 we write

\begin{align}
    Q_{D}(s,a) &\le (\mathcal{T}^{*}Q_{D})(s,a) \\
    \nonumber &= \sum_{s'} T(s, a, s')\left[ r_{D}(s,a,s') + \max_{a'}Q_{D}(s',a') \right] \\
    &= -\sum_{s' \in \mathcal{S}_{D}}T(s, a, s') -\sum_{s' \in C'_{D}}T(s, a, s') + \sum_{s' \notin \mathcal{S}_{D}\cup C'_{D}}T(s, a, s')\left[r_{D}(s,a,s') +  \max_{a'}Q_{D}(s',a')\right] \label{eq:line:prop1} \\
    &= -P_{D}(s,a) -F_{D}(s,a) - \beta_D(s,a)  \label{eq:prop1_proof}
\end{align}
in which, $-\beta_{D}$ is the last term of \eqref{eq:line:prop1}. The reward of $\mathcal{M}_{D}$ is always zero unless at death terminal states where $r_{D}(s,a,s')=-1$. Hence, assumption 3 implies that $\beta_{D}(s,a)$ is always non-negative, regardless of how much $Q_{D}(s',a')$ is inaccurate. The rest of argument in Theorem 1 remains valid with $Q_{D}$ and $\beta_{D}$ replacing $Q^{*}_{D}$ and $M_{D}$.

\hfill $\blacksquare$

\raggedright\textbf{Remark 1.}\setlength{\parindent}{15pt} One setting that holds assumption 3 of Proposition 1 is in the tabular case where each $Q_{D}(s, a)$ is stored separately and under the assumption that all $(s,a)$ pairs are initialized with any non-positive number (naturally in $[-1, 0]$). In the general case involving non-tabular estimators, a practical way to assure that Assumption 3 of Proposition 1 holds is to clip all the values at $-1$ and $0$. 

\raggedright\textbf{Remark 2.}\setlength{\parindent}{15pt} There are certain cases that formally satisfy assumption 2. For example, the true value of \emph{any} policy (not necessarily optimal) satisfies this inequality \cite{bertsekas_neuro}. Another example is in the tabular setting when all values are initialized \emph{pessimistically} (e.g., at $-1$); however, pessimistic initialization may increase false positives because all unseen $(s,a)$ pairs will be inferred as dead-ends. In other cases, since $Q_{D}(s, a)$ is the convergence point of Bellman error, it is likely that for many state-treatment pairs assumption 2 holds. Nevertheless, one should note that this assumption needs further scrutiny and may not hold in general when function approximation is used. In particular, over-estimation issue (if exists for any state-treatment pair) will forfeit assumption 2. 

\raggedright\textbf{Remark 3.}\setlength{\parindent}{15pt} Proposition 1 implies that under certain assumptions, at each state only the value of treatments that lead to dead-end states w.p.1 has to be fully converged. Importantly, such values are independent of the values of other (non-dead-end) states, since according to Lemma 3 a dead-end's next state is also always either a dead-end or a death terminal state, regardless of the administered treatment. In an abstract way, it leaves out the necessity of learning the value for all the resulting trajectories from other treatments at the initial state as well as in the future resulting trajectories, which grow exponentially. Hence, at least in the tabular case with $-1$ value-initialization, learning the treatment avoidance method by securing the behavioral policy is an exponentially smaller problem than learning optimal policy (or optimal values), which advises for best treatments. 

\raggedright\textbf{Remark 4.}\setlength{\parindent}{15pt} Full convergence of values of dead-end states $\mathcal{S}_{D}$ to -1 in Assumption 1 can be relaxed to $-(1-\epsilon)$ for some $\epsilon \in [0,1)$. In that case, rewriting \eqref{eq:prop1_proof} induces that the security guarantee will degrade to $\pi(s,a) \le 1 - (1-\epsilon)\lambda$. That is, for a risky treatment, abiding by $1+Q_{D}$ guarantees less decrease of its probability than what the security conditions requires. This may be addressed by adjusting the thresholds more conservatively. 

\clearpage

\section{Further Remarks on Related Work}
\label{appendix:sec:related}
In light of discussions with reviewers during the rebuttal period, we feel the need to honor similarities and differences between our work and those introduced in \citet{irpan2019off} more thoroughly than space constraints allow in the main body of this paper. While there are partial parallels in terms of grounding ideas, our theoretical development vastly diverges from~\citet{irpan2019off}, which relies wholly on empirical exploration and is centered wholly on policy evaluation rather than the assessment of specific decisions an agent may make.
We summarize key important differences as follows:
\begin{itemize}
\item Their concept of \textit{feasible} is simply being non-catastrophic and is different from \textit{rescue}, which is a state where recovery is reachable w.p.1. (i.e., there is no parallel for rescue states in their work).
\item The properties of the Q function and how it formally links to the probabilities of a state being feasible or catastrophic is not derived, discussed, or used in their work.
\item Their OPC metric is a proxy for evaluation/ranking learned policies. They do not use the framing to identify problematic or high-risk actions that may lead to catastrophic behavior. More accurately, there is no particular parallel for the concept of (treatment) security, its definition, and the formal guarantees which then shape the foundation of DeD.
\item In their work, the classification component is used to identify the value of state-action pairs on a binary $\{0, 1\}$ scale. This makes negative behavior somewhat unidentifiable (they acknowledge this) from intermediate feasible states that do not correspond to terminal conditions.
\item Our dead-end construction (reward of -1 for bad outcomes + no-discounting) provides an inherently different value function, which (with a negative sign) formally gives rise to the minimum probability of bad outcomes in the future.
\item Side note: in dangerous and stochastic environments and for sufficiently long episodes, their Theorem 1 results in the trivial bound  (since the lower-bound becomes a negative value). Their experiments are restricted to robotic tasks and the Atari game of Pong; thus, this core problem has remained hidden in their work.
\end{itemize}

At a high level both \citet{irpan2019off} and our work exploit constructed asymmetries within the state space to identify regions that are undesirable and should be avoided. The notions of \emph{feasible} and \emph{catastrophic} in \citet{irpan2019off} are related, in context of an optimal policy $\pi^*$, with $P_{\pi^*}(\text{success}|\text{feasible})>0$ where $P_{\pi^*}(\text{success}|\text{catastrophic})=0$ always. Thus, by being able to classify which states are \emph{catastrophic} evaluation of any trajectory containing such states is made significantly easier when evaluating policies developed from observational data. Irpan, et al. worked to label all state-action pairs as either \emph{feasible} or \emph{catastrophic} using positively-unlabeled classification. 

With a similar asymmetry, but generalized to encompass the delicate dynamics often observed in safety-critical domains, we formalize the relationship between the special states (described in Section~\ref{sec:special_states}) and the terminal conditions of death or recovery as follows: $P(\text{recovery}|\text{rescue}) = 1$ for some policy $\pi$ (including the optimal policy $\pi^*$). In contrast, dead-end states have a more extreme condition where $P(\text{death}|\text{dead-end}) = 1$ for all policies $\pi$. This helps to emphasize the importance of identifying treatments that may lead to dead-end states and subsequently influence decision-makers to avoid selecting those treatments. The means by which we infer the risk of a treatment (or action) is through a pair of independent MDPs used to identify the value of a state-treatment pair in accordance to its risk of being a dead-end or the chance it may lead to rescue and being a rescue state. This joint inference problem is used to affix and confirm whether a state should be avoided (and all treatments leading to this state) as discussed in Theorem 1 in Section~\ref{sec:method}.

\clearpage

\section{Toy Problem: LifeGate}
\label{sec:lifegate_apdx}

\begin{figure}[t!]
    \centering
    \includegraphics[width=0.8\textwidth]{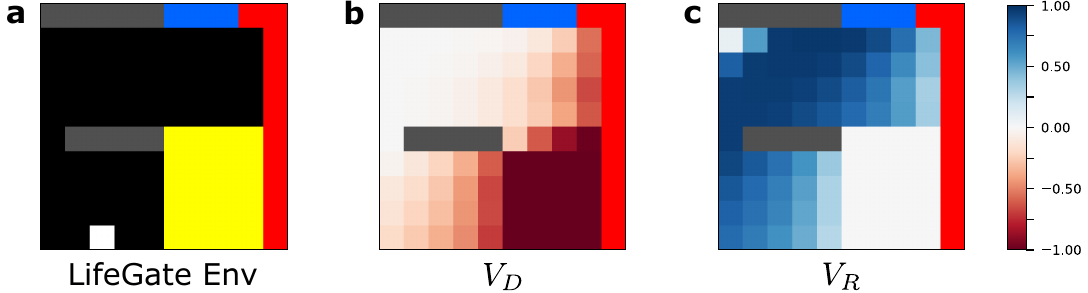}
    \caption{\textbf{The Life-Gate Example.} The tabular navigation task of life-gate is illustrated in (\textbf{a}). Its corresponding dead-end and rescue (nearly optimal) state-value functions, $V^{*}_{D}$ and $V^{*}_{R}$, ares shown in (\textbf{b}) and (\textbf{c}), respectively. The value functions are learned through Q-learning and with the definition of $\mathcal{M}_D$ and $\mathcal{M}_R$.}
    \label{fig:lifegate_exp}
\end{figure}

In this section we present a detailed toy-example with tabular state-space, called Life-Gate (\ref{fig:lifegate_exp}). The white square depicts the agent's position, which has five actions corresponding to moving up, down, left, right, and doing nothing (no-up). The grey walls are neutral barriers, hitting to which does not have any effect, yet the agent cannot pass them. There are two possible terminal states: (1) death gates (shown in red) and (2) life gates (shown in blue), and the goal is for the agent to reach a life gate. If the agent lies in black areas, any action will cause a forceful move to the right with $\textsc{death-drift}=40\%$ probability, or otherwise performs a cardinal move as expected. On the other hand, the yellow areas are all \textit{dead-end} states. If the agent reaches any of the yellow positions, at each step afterwards, the agent will move to the right with $70\%$ probability or remain put with $30\%$ probability, regardless of the taken action. Hence, the agent will be on an inescapable dead-end trajectory to a death gate with random length. However, the agent will not see any of the colors and the state only comprises agent's x-y position. 

We use Q-learning to compute the value functions of $\mathcal{M}_D$ and $\mathcal{M}_R$ as detailed in the main text: using discount of $\gamma=1$ for both, and $\mathcal{M}_R$ only assigns the reward of $+1$ in the case of reaching a life gate (zero otherwise), while $\mathcal{M}_D$ assign the reward of $-1$ if transitioning to a death gate (and zero otherwise). We stop training before full convergence; hence, there are possible learning errors (e.g., upper left corner for $V^*_R$). Of note, the value of walls (which are all zero) is simply an artifact of choosing zero for initialization of the Q-tables.

This set-up comprises an interesting case. The agent faces an environment to explore, with no knowledge of possible dangers. Importantly, once a dead-end state is reached, it may take some random number of steps before reaching a death gate, where the agent would realise expiration. All along such trajectories of dead-end states, the agent still has to choose actions with the (false) hope of reaching a life-gate. Discovering any single dead-end state and signaling the agent when the state gets in the scope would be of significant importance. On the other hand, the adjacent states to dead-ends are possibly the most critical ones to alert, as it might be the last chance to still do something. 

Let us probe this problem with the tools provided by Theorem 1. Based on Theorem 1, there are thresholds $\delta_D$ and $\delta_R$ which completely separate the values of dead-end states from the rest, both in term of $V^*_D$ and $V^*_R$. In this example, even with the errors due to lack of full convergence, $\delta_D = -0.7$ and $\delta_R = 0.7$ seem to clearly set the boundary for most states. There are some exceptions though. For example the top right corner is a false positive for $V^*_R$ due to learning errors, or at the top row of the yellow area, all the states are dead-ends but not all of them passes this test, again due to learning errors. If a state is observed whose $V_D$ and $V_R$ values violate these thresholds, the state can be flagged as a dead-end with high probability. Setting a lower threshold can help to raise flags earlier on, when the conditions are becoming high-risk, but it is not too late. We can see that $\delta_D = -0.2$ can act as a early-warning flag. Lastly, to also see (T1) and (T2) of Theorem 1, we note that only for all the yellow area (setting aside the few erroneous states), $V_D=-1$ and $V_R=0$.

\clearpage

\section{State Construction Details and Training} \label{sec:state_const}

\begin{figure*}[t!]
\centering
\includegraphics[clip, trim={0 0 0 0}, width={5.5in}]{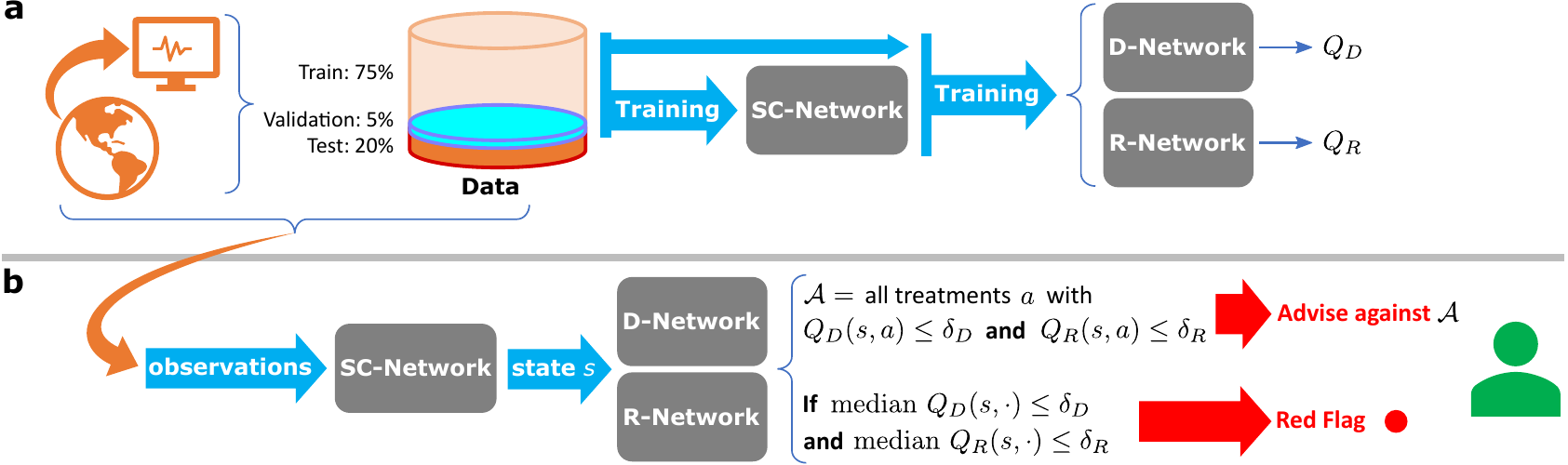}
\caption{\textbf{Dead-ends Discovery (DeD)}. Our pipeline includes two phases: (\textbf{a}) Training phase: using real-world data, we train the three neural networks to set-up i) state construction (SC-Network), ii) dead-end values (D-Network) and iii) rescue values (R-Network). (\textbf{b}) Test phase: the trained networks are used to map the immediate history of observations and the last action into $Q_{D}$ and $Q_{R}$ to infer risky conditions and dead-end outcomes, which is passed to the human decision-maker.
}
\label{fig:process}
\end{figure*}

This section highlights the construction and development of the state construction network used to embed the observation sequences of a patient's health condition into a state representation to be used in the reinforcement learning networks used for the detection and avoidance of dead-end states.

\subsection{Notation} 
Let {\small $\mathcal{D} = \{\tau_j\}_{j=1}^{n}$} denote the batch data of $n$ trajectories obtained from the database of patients with sepsis in the intensive care usit. We assume that this data is generated from a time-homogeneous partially observable Markov decision process (POMDP). Each trajectory $\tau_j$ has a finite number of transitions $m_j$. 
Each transition in a trajectory $j$ is a tuple with four entries {\small $(O_{t,j}, A_{t,j}, R_{t,j}, O_{t+1,j})$}, where {\small $j \in \{1, \dots, n\}$}, {\small $t \in \{1, \dots, m_j\}$}. 
The observation and action (treatment) spaces are defined as in Sec.~\ref{sec:data} where:
\begin{itemize}
    \item $O_{t,j}, O_{t+1, j} \in \mathcal{O}$ are the observations received at times $t$ and $t+1$ respectively in trajectory $j$ and $\mathcal{O} \subset \mathbb{R}^{d_{\mathcal{O}}}$ is the observation space. In our case for the sepsis treatment problem, the observation space is 44 dimensional.
    \item $A_{t,j} \in \mathcal{A}$ is the action taken at time $t$ in trajectory $j$ and $\mathcal{A}$ is the action space. In this work we restrict attention to discrete action spaces of finite cardinality, $|\mathcal{A}| = n_a$. In our case for the sepsis treatment problem, $n_a = 25$.
    \item $R_{t,j} \in \mathbb{R}$ is the per-step reward received at time $k$ in trajectory $j$. We use an end-of-trajectory binary reward signal of $\pm 1$ (we, however, do not explicitly make use of the reward in the state construction network because we only focus on state representation learning for dynamics prediction).
\end{itemize}

For clarity we drop the trajectory index $j$ throughout the remainder of this section unless it is necessary to differentiate between trajectories. Let {\small $\hat{d}_{\mathcal{S}}$} denote the dimension of the learned state representation ($\hat{S}$), which is a hyper-parameter that needs to be chosen. 
Our objective is to learn a state construction function {\small $\psi: \{O_{0:t}, A_{0:t-1}\} \mapsto \hat{S}_{t},~t \ge 1$}, and {\small $\hat{S}_{t} \in \hat{\mathcal{S}} \subset \mathbb{R}^{\hat{d}_{\mathcal{S}}}$}. 
In addition to $\psi$, the approaches outlined in the next section also involve another function: a dynamics predictor $\phi$ that involves predicting the next observation $\hat{O}_{t+1}$. 
Hence, the function {\small $\phi: \hat{\mathcal{S}} \times \mathcal{A} \to \Delta(\mathcal{O})$}, where $\Delta(x)$ denotes a probability distribution of $x$, estimates the conditional distribution of the next observation given the current state representation and action. 

\subsection{State Construction (SC) Network}

We construct the state representation of a patient's condition by training a set of coupled functions, as motivated by the Approximate Information State (AIS) approach~\cite{subramanian2019approximate}. AIS satisfies two key properties: 1) each state is ``Markovian'' or sufficient for the prediction of the next state, and 2) observations are distinguishable when mapped to their corresponding states if they result in different future trajectories. The first function, denoted by $\psi$, encodes the observed sequence patient conditions and the treatments administered into a compressed representation. This representation (corresponding to the state used in the reinforcement learning networks) is then passed, along with the current treatment, to a decoding function $\phi$ to predict the next patient observation.

\begin{figure}[h]
\centering
\includegraphics[width=0.45\textwidth]{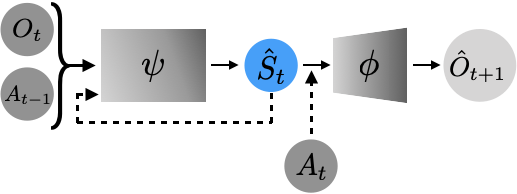}
\caption{The state construction network, comprised of the encoding function $\psi$ that provides the state representation $\hat{S}_t$ that is used with the decoding function $\phi$ to predict the next observation $\hat{O}_{t+1}$.}
\label{fig:ais_arch}
\end{figure}

The input to $\psi$ is the concatenation of the observation $O_t$ and last selected action $A_{t-1}$. For the function $\psi$ we use a 3-layer Recurrent Neural Network (RNN), where the first layer is a fully connected layer that maps the current observation and action (69 dimensional input: 44 dimensional observation with a 25 dimensional one-hot encoded action) to 64 units with ReLU activation.
This is followed by another $(64,128)$ fully connected layer with ReLU activation which is followed by a gated recurrent unit~\cite{cho2014properties} layer with hidden state size $\hat{d}_{\mathcal{S}}$. 
The output of this recurrent layer is used as the state representation $\hat{S}_t$.
The current action $A_t$ is concatenated to the state representation $\hat{S}_t$ and then fed through the decoder function $\phi$ to predict the next observation $\hat{O}_{t+1}$. The function $\phi$ is comprised of a three layer neural network with sizes $(\hat{d}_{\mathcal{S}}+25,64)$, $(64,128)$ and $(128,44)$ (with ReLU activation for the first two layers). The last layer outputs a 44-dimensional vector, which forms the mean vector of a unit-variance multivariate Gaussian distribution, samples from which are used to predict the next observation. A schematic of the the state construction network is provided in \ref{fig:ais_arch}. The two functions $\psi$ and $\phi$ that comprise the state construction network are jointly trained by maximizing the negative log likelihood of the predicted next observation $\hat{O}_{t+1}$. 

This is formulated by maximizing the objective:
\begin{align*}
    \mathcal{L}(\mathcal{O}_{t+1},\hat{\mathcal{O}}_{t+1}) = - \sum^{d_{\mathcal{O}_{j}}} \log\mathcal{N}(\mathcal{O}_{t+1,j}; \mu_j, \sigma_j^2)
\end{align*}
where $\mu_j = \hat{\mathcal{O}}_{t+1}, \sigma_j^2 = 1, \text{ and } \hat{\mathcal{O}}_{t+1}=\psi(~\phi(~O_t,~A_{t-1}), ~A_t)$.

\subsection{Hyperparameter selection} 
\begin{figure}[h]
\centering
\includegraphics[width=0.4\textwidth]{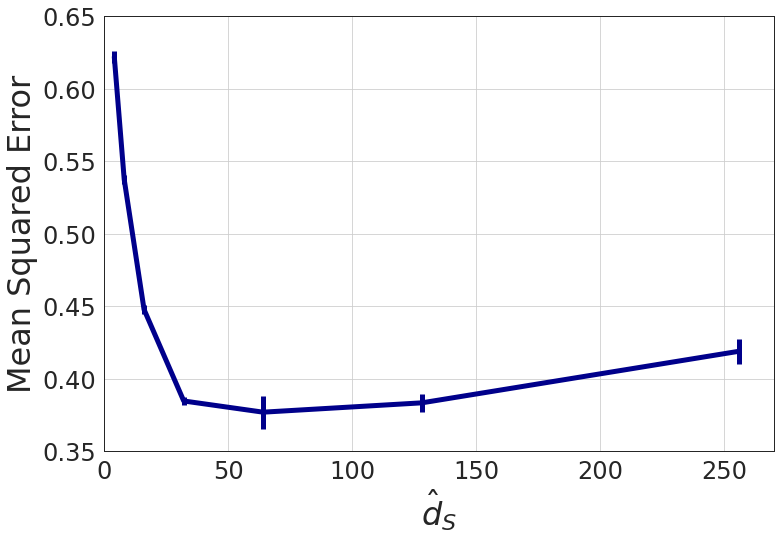}
\caption{Analysis of setting the dimension of the learned state representation $\hat{d}_{\mathcal{S}}$ and its effect on the accuracy of predicting the next observation. The bars represent standard deviation. With this, we determine to set $\hat{d}_{\mathcal{S}}=64$ in the SC-network.}
\label{fig:ais_eval}
\end{figure}

The dimension of the state representation $\hat{d}_{\mathcal{S}}$ was chosen from among $\{4, 8, 16, 32, 64, 128, 256\}$ dimensions. The choices of the size of neural network layers was chosen proportional to the size of $\hat{d}_{\mathcal{S}}$, with the final values reported in the prior subsection following the optimal choice of $\hat{d}_{\mathcal{S}}$ being equal to 64. The model construction network was trained for 600 epochs with learning rates of $\{0.0001,0.0005,0.001,0.005,0.001\}$ with the choice of $lr~=~0.0005$ providing the optimal training of the network. We demonstrate the evaluation of the choice of the dimension for the state representation in \ref{fig:ais_eval}.

\clearpage

\section{D- and R-Networks Training Details} \label{appendix:sec:rl-training}

We use double-DQN algorithm to train both networks. We refer the reader to our code for the implementation details (and we tried to make the code straightforward and relatively easy to understand). In particular, both D- and R-Networks consist of two linear layers with 64 nodes. The first layer is followed by ReLU nonlinearity and the second layer directly outputs 25 nodes corresponding to the 25 treatments. We use learning rate of $0.0001$ and minibatch size of $64$. In each minibatch, we select 62 transitions uniformly from the train data and append it with two uniformly selected ``death'' transitions (last transitions of nonsurvivor patients). All other chosen hyper-parameters can be found in the \emph{config.yaml} file in the root directory of our code.

\clearpage

\section{Data Details} \label{sec:data}

We use the MIMIC (Medical Information Mart for Intensive Care) - III dataset (v1.4), which has been sourced from the Beth Israel Deaconess Medical Center in Boston, Massachusetts~\cite{JohnsonPollardShenEtAl2016,mimicweb}. This dataset comprises of deidentified patient treatment records of patients admitted to critical care units (CCU, CSRU, MICU, SICU, TSICU). The database includes data collected from 53,423 distinct hospital admissions of patients over 16 years of age for a period of 12 years from 2001 to 2012. The MIMIC dataset has been used in many reinforcement learning for health care projects, including mechanical ventilation and sepsis treatment problems. There are various preprocessing steps that are performed on the MIMIC-III dataset in order to obtain the cohort of patients and their relevant observables for the sepsis treatment study. 

To extract and process the data, we follow the approach described in~\cite{komorowski2018artificial} and the associated code repository given in~\cite{aiclinician}. This includes all ICU patients over 18 years of age who have some presumed onset of sepsis (following the Sepsis 3 criterion) during their initial encounter in the ICU after admission, with a duration of at least 12 hours. These criteria provide a cohort of 19,611 patients, among which there is an observed mortality rate just above 9\%, where mortality is determined by patient expiration within 48h of the final observation.  Observations are processed and aggregated into 4h windows with treatment decisions (administering fluids, vasopressors, or both) discretized into 5 volumetric categories. All data is normalized to zero-mean and unit variance and missing values are imputed using k-Nearest Neighbor imputation, where possible. In the absence of similar observations any remaining missing values filled with the population mean. We report the 44 features used for the Dead-end approach proposed in this paper in \ref{tab:features} with high-level statistics for the extracted cohort in \ref{tab:cohort_statistics}.

\begin{table}[!ht]
\caption{Patient features used for learning state representations for predicting future observations}
% \vspace{-0.3cm}
\label{tab:features}
\begin{center}
{\small
\begin{tabular}{l|l|l|l}
\toprule
Age & Gender & Weight (kg) & Re-admission \\
Glasgow Coma Scale & Heart Rate & Sys. BP & Dia. BP          \\
Mean BP & Respiratory Rate & Body Temp (C) & FiO2   \\
Potassium & Sodium & Chloride & Glucose \\
INR & Magnesium & Calcium & Hemoglobin \\
White Blood Cells & Platelets & PTT & PT\\
Arterial pH & Lactate & PaO2 & PaCO2\\
PaO2 / FiO2  & Bicarbonate (HCO3)  & SpO2 & BUN \\
Creatinine & SGOT & SGPT & Total Bilirubin \\
Output (4h) & Output (total) & Cumulated Balance & SOFA \\ 
SIRS & Shock Index & Base Excess & Mech. Ventilation \\
\bottomrule
\end{tabular}
}
\end{center}
\end{table}

\begin{table}[!ht]
\caption{MIMIC Sepsis Cohort Statistics}
\label{tab:cohort_statistics}
{\footnotesize
\begin{tabular}{llll}
\hline
                                          Variable &         MIMIC ($n = 19611$) & Variable & MIMIC ($n=19611$) \\
\hline
                             \textbf{Demographics} &    & \textbf{Outcomes}  &           \\
                                        Age, years &     66.2 (53.8-78.1)    & Deceased &          1881 (9.6\%) \\
                                  Age range, years &    & Vasopressors administered &         5664 (28.9\%) \\
                                \hspace{5mm}18-29  &  741 (3.8\%) & Fluids administered &         17812 (90.8\%) \\
                                \hspace{5mm}30-39 & 896 (4.6\%) & Ventilator used &         9353 (47.7\%) \\
                                \hspace{5mm}40-49 & 2029 (10.3\%)    & &  \\
                                \hspace{5mm}50-59 & 3471 (17.7\%)  & \textbf{Severity Scores} & \\
                                \hspace{5mm}60-69 &         4321 (22.0\%)  & SOFA & 5 (3.0-8.0) \\
                                \hspace{5mm}70-79 &         4086 (20.8\%)  & SIRS & 2 (1.0-2.0) \\
                                \hspace{5mm}80-89 &         3069 (15.6\%)  & Shock Index & 0.72 (0.6-0.86) \\
                                \hspace{5mm}$\geq$90 &         998 (5.1\%)   & &  \\
                                            Gender &                     & & \\
                                  \hspace{5mm}Male &         10917 (55.6\%) & & \\
                                \hspace{5mm}Female &        8694 (44.3\%) & & \\
                                Re-admissions      &    1424 (7.3\%) & & \vspace{0.5em}\\
                  \textbf{Physical exam findings} &                 & &      \\
                          Temperature ($^{\circ}$C) &     37.2 (36.6-37.7) & & \\
                          Weight (kg) &     79.7 (66.7-95.2) & & \\
                     Heart rate (beats per minute) &      86.0 (75.0-98.0) & & \\
             Respiratory rate (breaths per minute) &      19.8 (16.6-23.3) & & \\
                    Systolic blood pressure (mmHg) &   118.3 (105.8-133.6) & & \\
                  Diastolic blood pressure (mmHg) &     56.6 (48.6-65.4) & &\\
                     Mean arterial pressure (mmHg) &     77.0 (69.0-86.7) & & \\
                     Fraction of inspired oxygen (\%) &  40.0 (35.0-50.0) & & \\
                     P/F ratio                     &     307.5 (192.0-579.0) & &\\
                                Glasgow Coma Scale &      14.8 (11.0-15.0) & & \vspace{0.5em}\\
                      \textbf{Laboratory findings} &     & & \\
                                        Hemotology & & Coagulation &     \\
  \hspace{5mm}White blood cells (thousands/$\mu$L) & 10.8 (7.7-14.8) & \hspace{5mm}Prothrombin time (sec) & 14.3 (13.1-16.4) \\
          \hspace{5mm}Platelets (thousands/$\mu$L) &  202.0 (137.0-286.0) & \hspace{5mm}Partial thromboplastin time (sec) & 32.6 (27.6-44.9) \\
                    \hspace{5mm}Hemoglobin (mg/dL) & 10.2 (9.1-11.4) & \hspace{5mm}INR &     1.3 (1.1-1.5) \\
                  \hspace{5mm}Base Excess (mmol/L) & 0.5 (0.0-2.6) & &  \\
                                         Chemistry &                      & Blood gas & \\
                      \hspace{5mm}Sodium (mmol/L) &  138.9 (136.0-141.0) & \hspace{5mm}pH &     7.41 (7.35-7.44) \\ 
                    \hspace{5mm}Potassium (mmol/L) & 4.0 (3.7-4.4) & \hspace{5mm}Oxygen saturation (\%) &     97.3 (95.5-98.8) \\ 
                     \hspace{5mm}Chloride (mmol/L) &  105.0 (101.0-108.5) & \hspace{5mm}Partial pressure of O2 (mmHg) &    124.0 (85.0-241.1) \\ 
                  \hspace{5mm}Bicarbonate (mmol/L) & 25.0 (22.0-28.0) & \hspace{5mm}Partial pressure of CO2 (mmHg) &     40.6 (36.0-46.0) \\
                        \hspace{5mm}Calcium (mg/L) & 8.3 (7.8-8.8) & & \\
                      \hspace{5mm}Magnesium (mg/L) & 2.0 (1.8-2.2) & & \\
          \hspace{5mm}Blood urea nitrogen (mg/dL) & 22.0 (14.0-36.0) & & \\
                    \hspace{5mm}Creatinine (mg/dL) & 1.0 (0.7-1.5) & & \\
                      \hspace{5mm}Glucose (mg/dL) & 127.4 (107.0-156.0) & & \\
                        \hspace{5mm}SGOT (units/L) & 38.0 (22.0-74.0) & & \\
                        \hspace{5mm}SGPT (units/L) & 30.0 (17.0-64.0) & &  \\
                        \hspace{5mm}Lactate (mg/L) & 1.5 (1.1-2.2) & & \\
                \hspace{5mm}Total bilirubin (mg/L) & 0.7 (0.4-1.5) & & \\
\hline
\end{tabular}
}
\end{table}

\clearpage

\section{Supporting Figures and Tables}

\vspace{1.5in}

\begin{table}[h]
\centering
\begin{tabular}[t]{@{} r | c c c c | c c c c | c c c c @{}}
\multicolumn{13}{l}{\textbf{{\large a}~~Red flag thresholds}} \\\toprule
& \multicolumn{4}{c}{\textbf{D-Network}} & \multicolumn{4}{c}{\textbf{R-Network}} & \multicolumn{4}{c}{\textbf{Full}} \\
& \multicolumn{2}{c}{Survivors} & \multicolumn{2}{c}{Nonsurvivors} & \multicolumn{2}{c}{Survivors} & 
\multicolumn{2}{c}{Nonsurvivors} & \multicolumn{2}{c}{Survivors} & \multicolumn{2}{c}{Nonsurvivors} \\
& $Q_{D}$ & $V_{D}$ & $Q_{D}$ & $V_{D}$ & $Q_{R}$ & $V_{R}$ & $Q_{R}$ & $V_{R}$ & $Q$ & $V$ & $Q$ & $V$ \\
\textbf{-72 h} & 0.2\% & 0.2\% & 0.0\% & 0.0\% & 0.5\% & 0.2\% & 2.8\% & 0.9\% & 0.0\% & 0.2\% & 0.0\% & 0.0\% \\
\textbf{-48 h} & 1.2\% & 0.4\% & 8.1\% & 5.4\% & 1.5\% & 0.5\% & 5.9\% & 4.3\% & 0.7\% & 0.2\% & 2.7\% & 2.7\% \\
\textbf{-24 h} & 1.1\% & 0.4\% & 16.3\% & 13.0\% & 1.2\% & 0.3\% & 16.7\% & 13.0\% & 0.6\% & 0.1\% & 12.2\% & 10.6\% \\
\textbf{-12 h} & 0.9\% & 0.4\% & 20.2\% & 18.2\% & 0.7\% & 0.3\% & 20.2\% & 17.4\% & 0.4\% & 0.2\% & 12.8\% & 14.7\% \\
\textbf{-8 h} & 1.0\% & 0.4\% & 24.5\% & 21.9\% & 0.9\% & 0.3\% & 19.3\% & 20.4\% & 0.6\% & 0.2\% & 13.4\% & 17.8\% \\
\textbf{-4 h} & 1.2\% & 0.5\% & 29.7\% & 26.4\% & 0.7\% & 0.5\% & 24.9\% & 22.7\% & 0.5\% & 0.3\% & 20.1\% & 22.0\% \\
\bottomrule
\multicolumn{13}{l}{} \\
\end{tabular}
\begin{tabular}[t]{@{} r | c c c c | c c c c | c c c c @{}}
\multicolumn{13}{l}{\textbf{{\large a}~~Yellow flag thresholds}} \\\toprule
& \multicolumn{4}{c}{\textbf{D-Network}} & \multicolumn{4}{c}{\textbf{R-Network}} & \multicolumn{4}{c}{\textbf{Full}} \\
& \multicolumn{2}{c}{Survivors} & \multicolumn{2}{c}{Nonsurvivors} & \multicolumn{2}{c}{Survivors} & 
\multicolumn{2}{c}{Nonsurvivors} & \multicolumn{2}{c}{Survivors} & \multicolumn{2}{c}{Nonsurvivors} \\
& $Q_{D}$ & $V_{D}$ & $Q_{D}$ & $V_{D}$ & $Q_{R}$ & $V_{R}$ & $Q_{R}$ & $V_{R}$ & $Q$ & $V$ & $Q$ & $V$ \\
\textbf{-72 h} & 1.6\% & 0.5\% & 4.6\% & 2.8\% & 1.8\% & 0.2\% & 5.6\% & 2.8\% & 0.5\% & 0.0\% & 2.8\% & 2.8\% \\
\textbf{-48 h} & 3.1\% & 2.2\% & 12.4\% & 11.9\% & 2.7\% & 2.1\% & 14.1\% & 11.9\% & 1.6\% & 1.5\% & 10.3\% & 9.7\% \\
\textbf{-24 h} & 2.7\% & 1.8\% & 17.1\% & 20.3\% & 2.2\% & 1.8\% & 13.8\% & 15.9\% & 1.4\% & 1.4\% & 10.6\% & 13.4\% \\
\textbf{-12 h} & 3.3\% & 2.5\% & 19.4\% & 19.4\% & 3.4\% & 2.4\% & 17.4\% & 17.8\% & 2.0\% & 1.7\% & 15.1\% & 17.1\% \\
\textbf{-8 h} & 3.0\% & 2.1\% & 20.8\% & 21.9\% & 2.6\% & 2.1\% & 21.6\% & 17.8\% & 1.2\% & 1.5\% & 18.6\% & 15.6\% \\
\textbf{-4 h} & 3.2\% & 2.4\% & 20.1\% & 20.1\% & 3.0\% & 2.4\% & 16.8\% & 21.2\% & 1.7\% & 1.5\% & 16.1\% & 17.6\% \\
\bottomrule
\end{tabular}
\vspace{0.2in}
\caption{\textbf{Prediction of potentially life-threatening treatments and states (full list).} Similarly to \ref{fig:hist_values}, the results correspond to the part of test data that satisfies having minimum length of the corresponding time step (X hours before terminal). To raise a flag, a patient must concurrently violate the corresponding thresholds, as specified in \ref{fig:hist_values}. 
$Q$ columns correspond to the value of actually selected treatments, while $V$ columns correspond to the median value of patients' state at the corresponding time.
}
\label{table:prediction}
\end{table}

\clearpage

\begin{figure*}[t]
\centering
\includegraphics[width=6.7in]{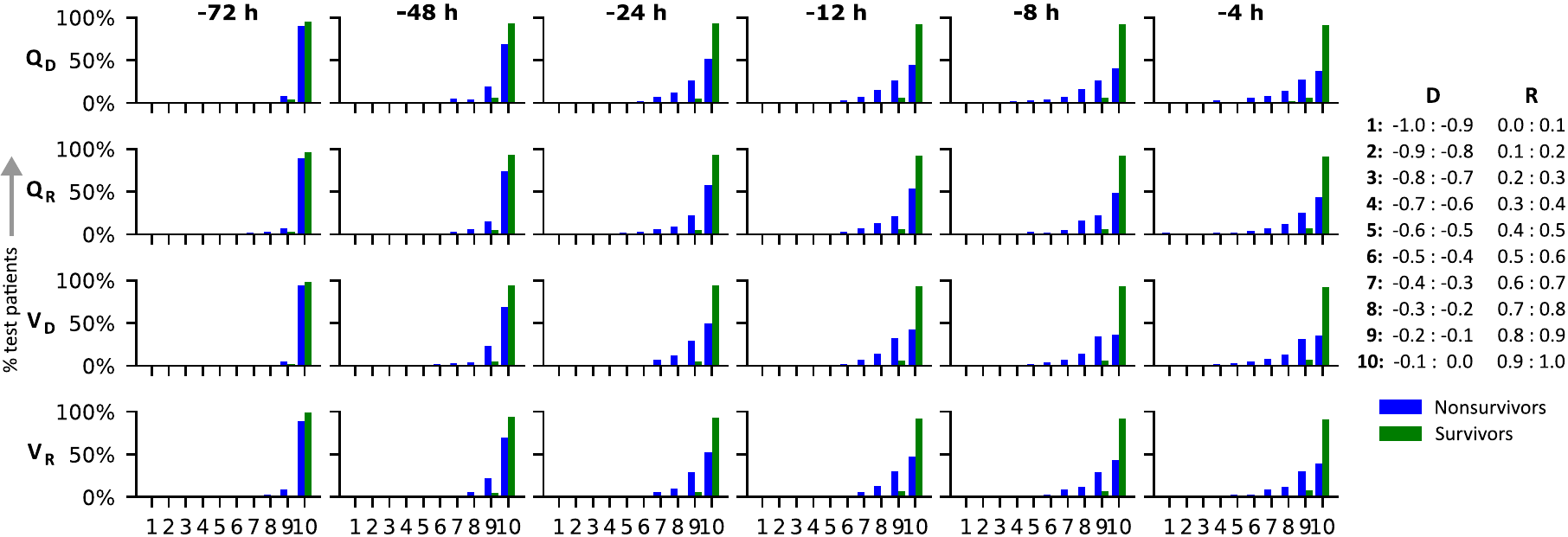}
\caption{\textbf{Full histogram of values in different time steps.} The histograms are plotted from the part of test data that satisfies having minimum length of the time step. The four rows are corresponding to the following: $V_{D}$ and $V_{R}$: median value of states from D-Network and R-Network, respectively, and $Q_{D}$ and $Q_{R}$: value of the selected treatments at the given time step from D-Network and R-Network, respectively. Note the distinctive difference between the trend of values in survivor (green bars) and nonsurvivor (navy bars) trajectories. In particular, in the course of 72 hours in the ICU, there is not much change in the value of selected or median treatment for the survivor patients, which is completely in contrast with those of nonsurvivor patients. 
}
\label{fig:hist_full_values}
\end{figure*}

\begin{figure*}[t]
\centering
\includegraphics{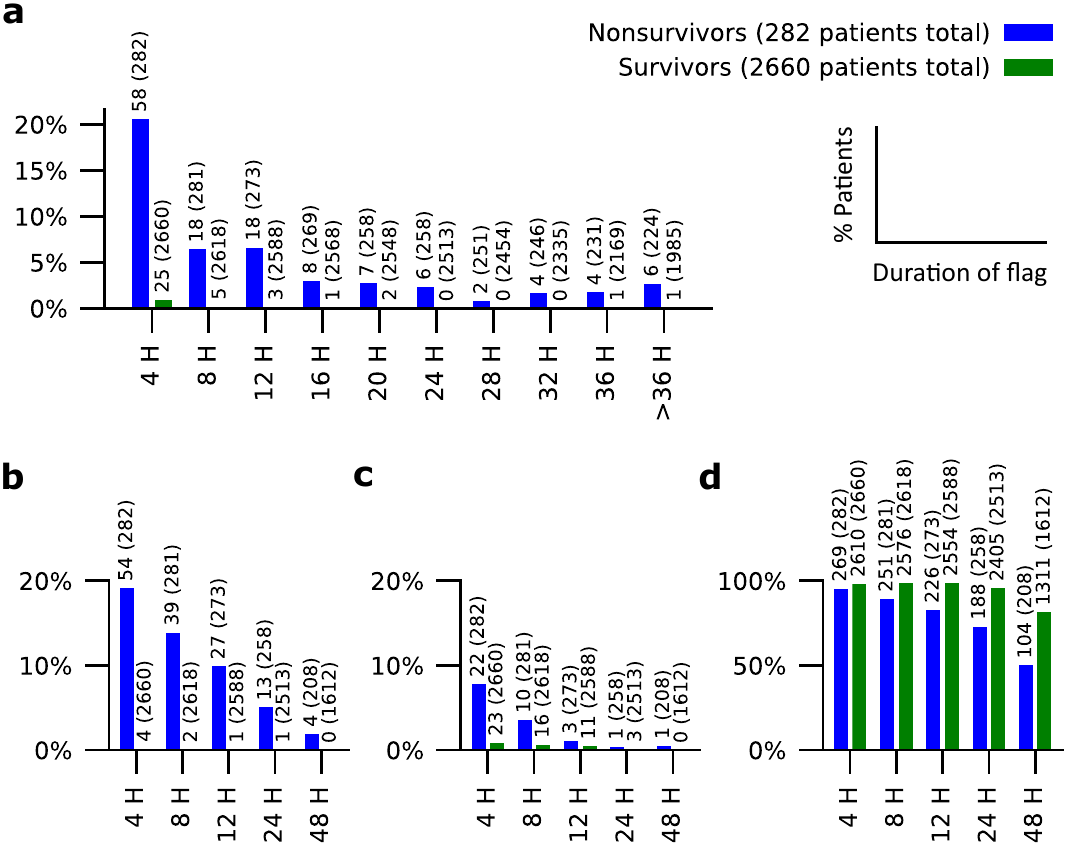}
\caption{\textbf{Flag duration for ICU patients.} Remaining on \emph{confirmed red-flag} is measured for both survivor and nonsurvivor patients. \textbf{a} The bars represent the percentage of patients who experience at least one red-flag with the exact duration on the horizontal axis. Texts depict number of patients (out of total patients with the minimum of specified stay duration). \textbf{b} and \textbf{c} depict patients who ``finish'' their ICU stay remaining on red and yellow flags, respectively, at the final X hours before terminal. \textbf{d} presents patients who ``start'' their trajectory with \emph{no flag} at all for the first X hours on the horizontal axis. We found that for the large part, both survivors and nonsurvivors start their trajectory without any flag, suggesting that they do not necessarily start with an unrecoverable situation. Further, nearly zero percent of survivors would raise and remain on red-flag for more than eight hours (even eight hours is quite rare compared to the total number of survivor patients). In contrast, nonsurvivor patients demonstrate a fat tail in the duration distribution \textbf{a} and repeatedly remain on the red-flag for eight hours or more. This result suggests that remaining on the red-flag for long periods strongly correlates with mortality, which is inline with our theoretical analysis.  
}
\label{fig:duration}
\end{figure*}

\begin{figure*}[t]
\centering
\includegraphics{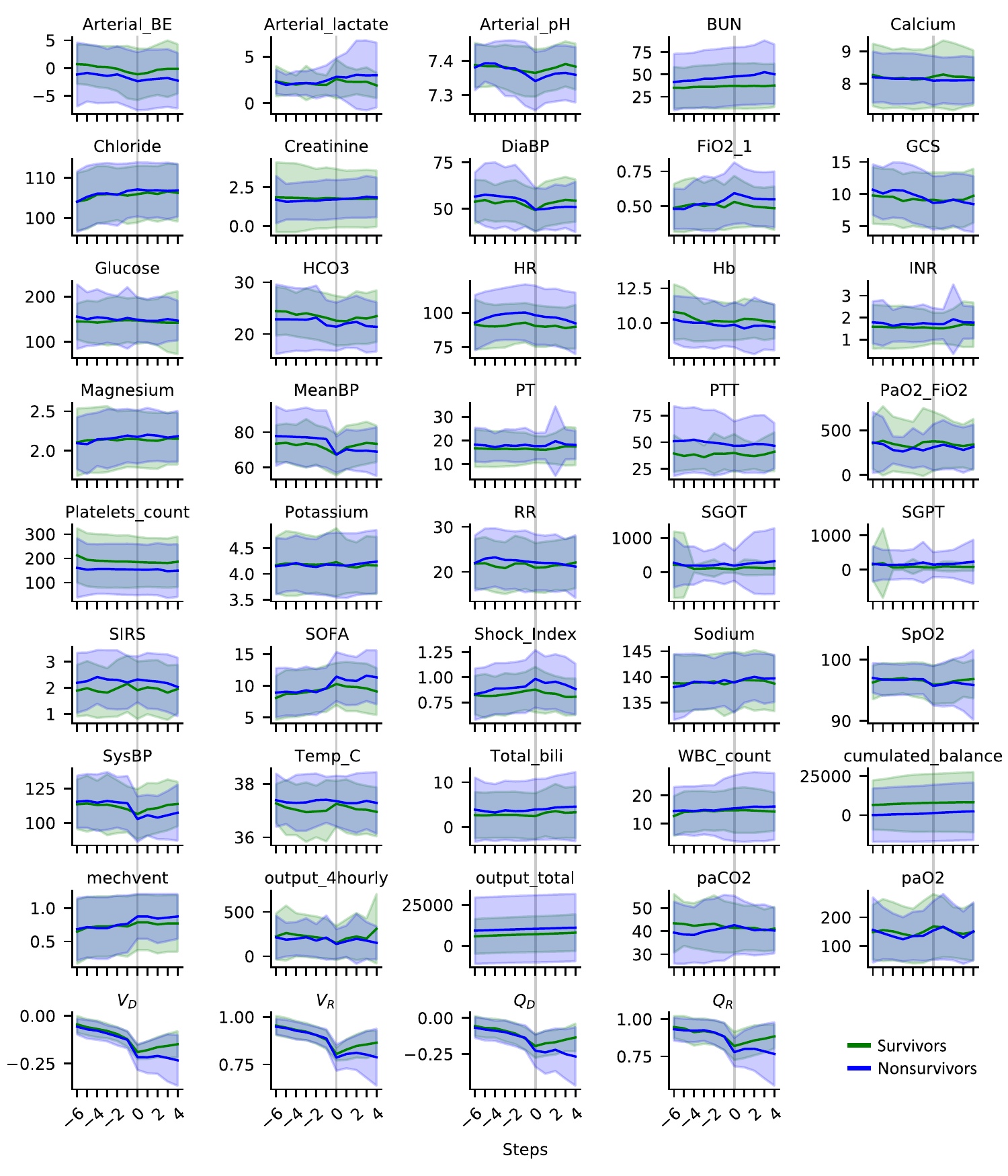}
\caption{\textbf{Signals prior to the first flag.} Complete list of vitals and standard measures in addition to our dead-end and secure values are shown for both survivor and nonsurvivor patients 24 hours (6 steps, 4 hours each) before and 16 hours (4 steps) after the first raised flag (red or yellow), indicated at point zero. Shaded areas represent standard deviation.
}
\label{fig:full-vitals}
\end{figure*}

\begin{figure*}[t!]
\centering
\includegraphics[width=6.7in]{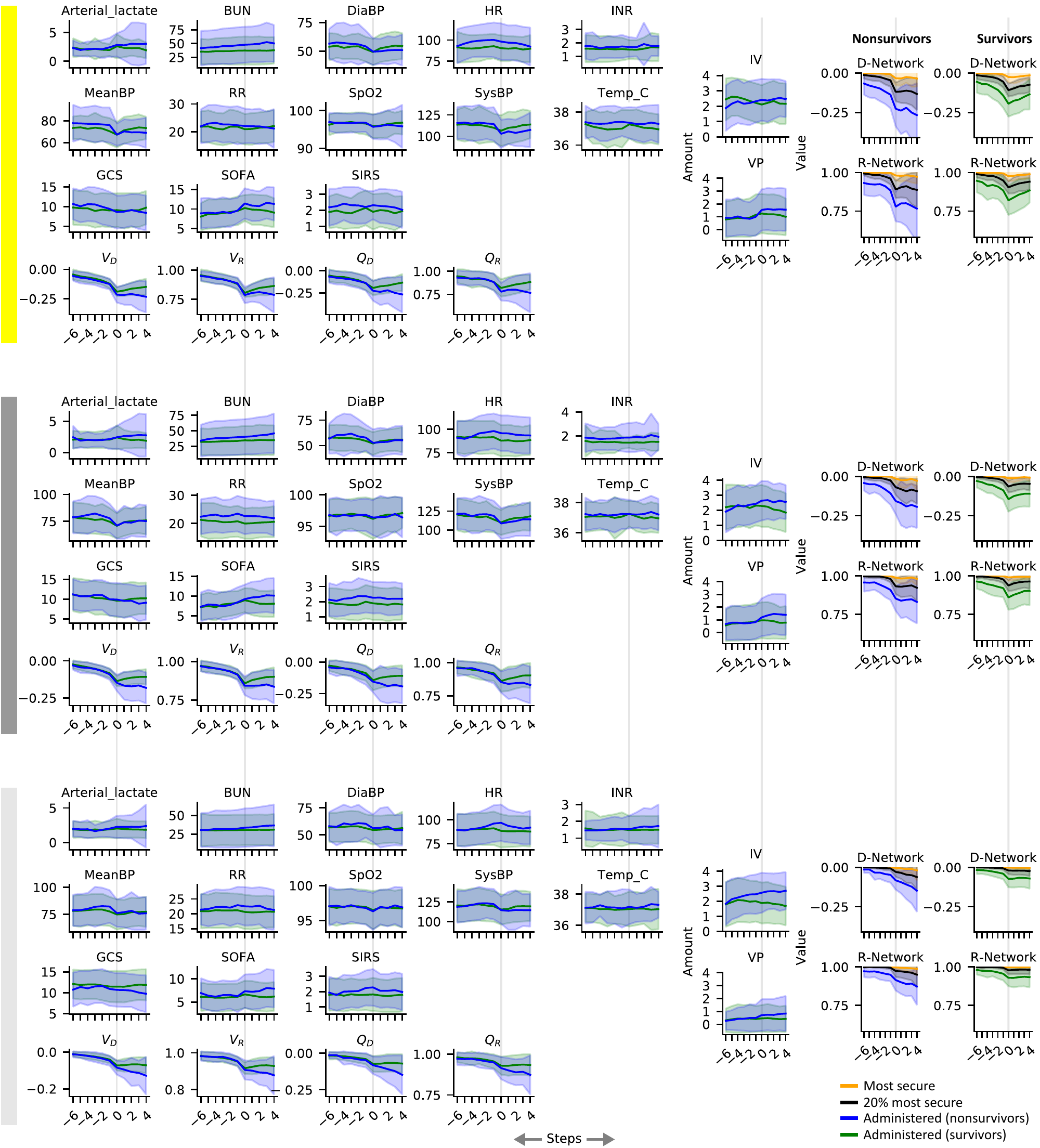}
\vspace{-0.25cm}
\caption{\textbf{Trend of various measures before and after the first raised flags.} Various measures are shown 24 hours (6 steps) before and 16 hours (4 steps) after the first threshold crossing. The colors respectively corresponds to the following thresholds: yellow: $\delta_{D}= -0.15$, $\delta_{R}= 0.85$; dark grey: $\delta_{D}= -0.10$, $\delta_{R}=0.90$; light grey: $\delta_{D}=-0.05$, $\delta_{R}=0.95$. Shaded areas represent standard deviation. 
}
\label{fig:measures_3flags}
\end{figure*}

\begin{figure*}[!ht]
\centering
\includegraphics[width=6.7in]{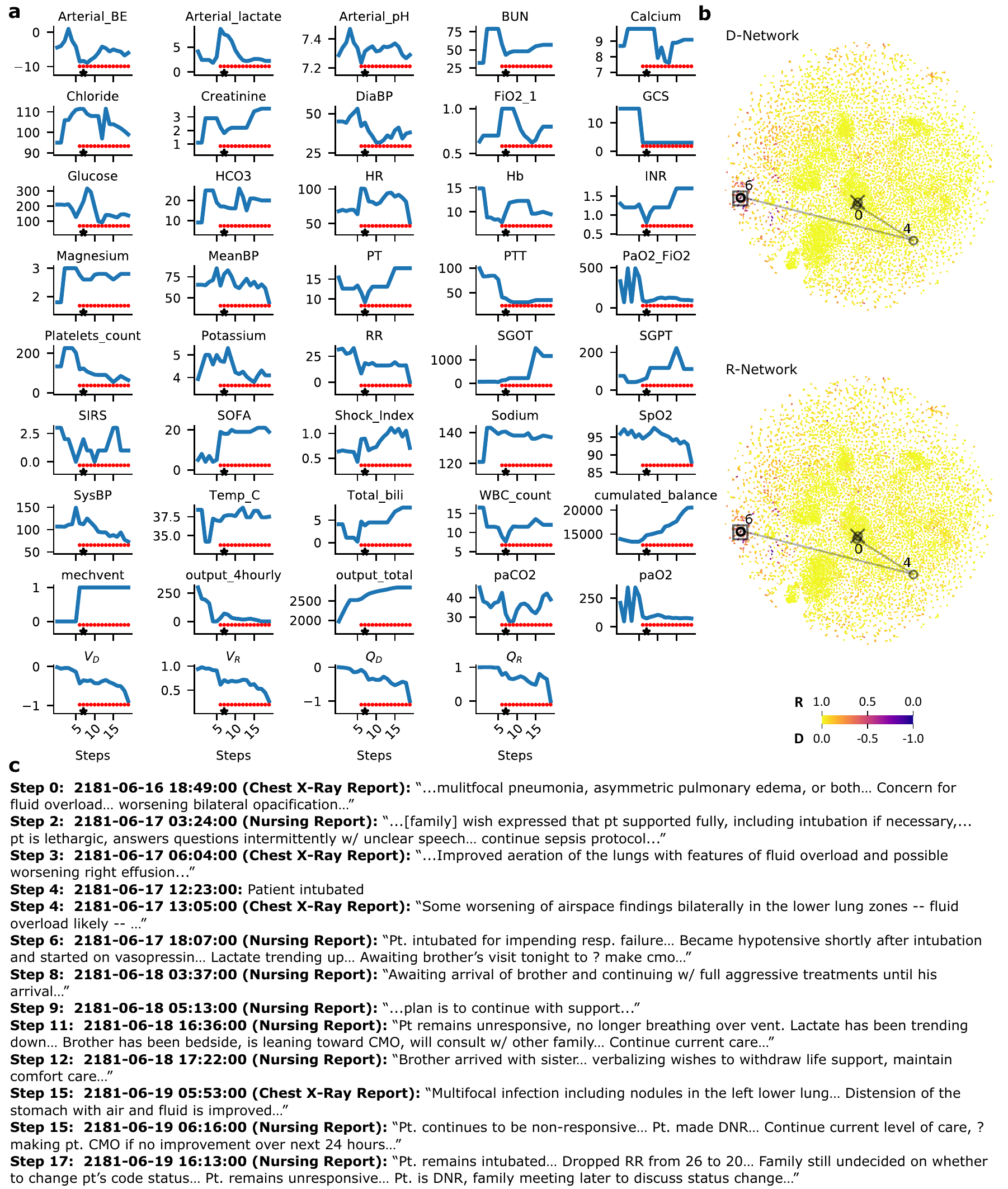}
\caption{\textbf{Complete analysis of nonsurvivor patient 262011.} \textbf{a} all the vitals, standard measures, max treatments, and network values for a nonsurvivor patient ICU-Stay-ID 262011. Red dots, yellow dots, and the asterisks show red and yellow flags and the presumed onset of sepsis, respectively. \textbf{b} patient's trajectory on the t-SNE plot, and \textbf{c} extracted chart notes from different source with their corresponding time stamp and quantized step.
}
\label{fig:full-traj_12139_icustayid_262011}
\end{figure*}

\begin{figure*}[!ht]
\centering
\includegraphics[width=6.7in]{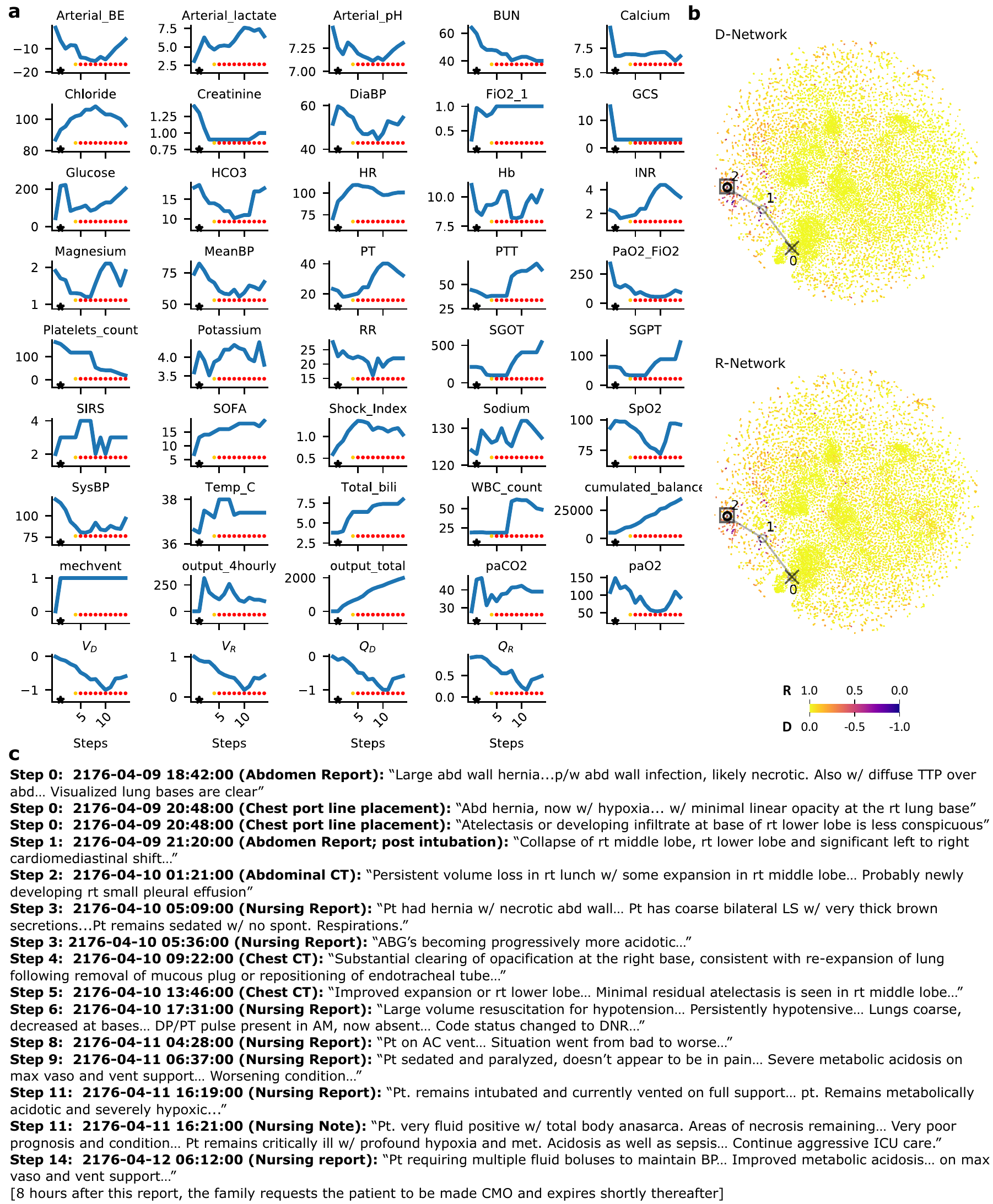}
\caption{\textbf{Complete analysis of nonsurvivor patient 270174.} \textbf{a} all the vitals, standard measures, max treatments, and network values for a nonsurvivor patient ICU-Stay-ID 270174. Red dots, yellow dots, and the asterisks show red and yellow flags and the presumed onset of sepsis, respectively. \textbf{b} patient's trajectory on the t-SNE plot, and \textbf{c} extracted chart notes from different source with their corresponding time stamp and quantized step.
}
\label{fig:full-traj_13778_icustayid_270174}
\end{figure*}

\begin{figure*}[!ht]
\centering
\includegraphics[width=6.7in]{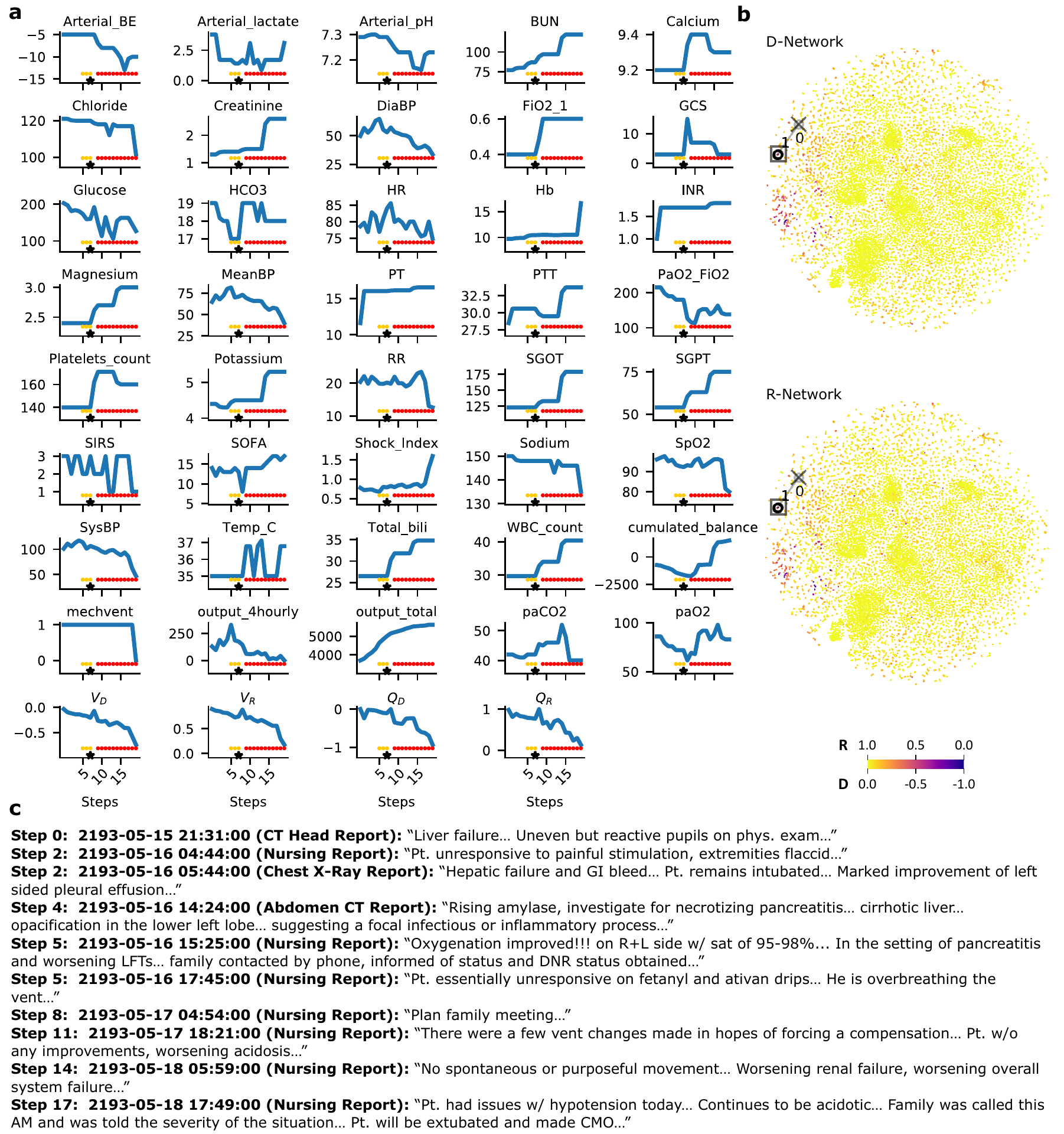}
\caption{\textbf{Complete analysis of nonsurvivor patient 235403.} \textbf{a} all the vitals, standard measures, max treatments, and network values for a nonsurvivor patient ICU-Stay-ID 235403. Red dots, yellow dots, and the asterisks show red and yellow flags and the presumed onset of sepsis, respectively. \textbf{b} patient's trajectory on the t-SNE plot, and \textbf{c} extracted chart notes from different source with their corresponding time stamp and quantized step.
}
\label{fig:full-traj_6938_icustayid_235403}
\end{figure*}

\end{document}